\documentclass[gray]{jmlr}

\usepackage{longtable}
\usepackage{graphicx}
\usepackage{url}
\usepackage{algorithm2e}
\usepackage{epstopdf}

\usepackage{comment}
\usepackage[normalem]{ulem}

\usepackage{booktabs}

\makeatletter
\let\Ginclude@graphics\@org@Ginclude@graphics 
\makeatother

\title[Augment to Interpret]{Augment to Interpret: Unsupervised and Inherently Interpretable Graph Embeddings}

\author{
    \Name{Gregory Scafarto} \Email{gregory.scafarto@euranova.eu}\\
    \Name{Madalina Ciortan} \Email{madalina.ciortan@euranova.eu}\\
    \Name{Simon Tihon} \Email{simon.tihon@euranova.eu}\\
    \Name{Quentin Ferré} \Email{quentin.ferre@euranova.eu}\\
    \addr Euranova, Brussels, Belgium
}

\begin{document}

\maketitle

\begin{abstract}

Unsupervised learning allows us to leverage unlabelled data, which has become abundantly available, and to create embeddings that are usable on a variety of downstream tasks. However, the typical lack of interpretability of unsupervised representation learning has become a limiting factor with regard to recent transparent-AI regulations.
In this paper, we study graph representation learning and we show that data augmentation that preserves semantics can be learned and used to produce interpretations. Our framework, which we named INGENIOUS, creates inherently interpretable embeddings and eliminates the need for costly additional post-hoc analysis. We also introduce additional metrics addressing the lack of formalism and metrics in the understudied area of unsupervised-representation-learning interpretability. Our results are supported by an experimental study applied to both graph-level and node-level tasks and show that interpretable embeddings provide state-of-the-art performance on subsequent downstream tasks.\footnote{Our code is available at \url{https://github.com/euranova/Augment_to_Interpret}.}

\end{abstract}


\section{Introduction}

Graphs are powerful data representations that model objects as well as the relationships between them.
Much like images, they are often complex to process.
A popular solution is to project such data into a latent space, thus creating vector embeddings, which can later be used by any classical machine learning training and inference pipeline.
These embeddings could be learned with supervision, but labelled data is often difficult to obtain.
This has resulted in an increased interest in the learning of semantic-preserving embeddings, which can later be used for a variety of downstream tasks.
There are numerous ways to learn graph representations~\citep{Representation_learning} without label supervision, but using graph neural networks (GNN) has become the go-to approach.
In particular, graph contrastive learning (GCL) aims at creating graph embeddings without supervision by leveraging augmented views of available samples to extract meaningful information.
This approach provided state-of-the-art results when applied to diverse domains ranging from chemistry~\citep{molecule_graph} to social sciences~\citep{social_science}.

\citet{simclr} demonstrated that the use of data augmentation in self-supervised contrastive learning can lead to the creation of semantic-preserving embeddings.
Similar results have been achieved on graph data by leveraging carefully crafted augmentation~\citep{GraphCL}.
Despite promising results, the underlying GNNs act as black-boxes. They are difficult to interpret, debug and trust, which raises the question: \textit{What input features do GNNs focus on when generating representations?}
Using edge and node dropping as graph augmentation can lead to the creation of sub-graphs that retain discriminative semantic information while reaching better results~\citep{GraphCL} than other augmentation schemes. Interestingly, this objective correlates with the aim of GNN interpretability, which is to identify highly influential sparse subsets of nodes and edges with the largest impact on the model's behaviour~\citep{PGexplainer}.
This observation leads us to the hypothesis that a well-learned augmentation can serve to produce interpretations of the information embedded in GNN representations.
Our main contributions follow.
\begin{itemize}
    \item We propose INGENIOUS (INherently INterpretable Graph and Node Unsupervised embeddings), a framework that generalises over existing approaches based on learned augmentation~\citep{adgcl,mega,GSAT} and we introduce new losses and a module to produce useful and interpretable embeddings.
    \item We show that embeddings produced by INGENIOUS are relevant to a variety of downstream tasks, matching or exceeding results obtained by state-of-the-art approaches.
    \item We show that carefully designed learned graph augmentation can be used to produce interpretations. We assess their quality along different axes introduced by \citet{interp_charac}:
    correctness, completeness, continuity, compactness and coherence. To this end, we introduce new metrics.
    \item We conduct a hyperparameter study, highlighting the importance of sparsity when using augmented views as interpretations.
\end{itemize}

To our knowledge, this is the first attempt at assessing the link between learned augmentation and interpretability. We characterise it on graph-level and node-level tasks. 
The structure of the paper is as follows:
We first position our work in Section~\ref{related_work} and present our framework in detail in Section~\ref{method}. In Section~\ref{experiments}, we assess the utility and interpretability of the framework by introducing necessary metrics and we perform an in-depth study of the properties of the framework. Finally, we discuss limitations and conclude.

\section{Related Work\label{related_work}}
Previous works on other modalities have shown the superiority of augmentation-based contrastive learning~\citep{simclr}.
However, adapting these works so that they work reliably on graph data raised new challenges, such as defining the right augmentation techniques amongst existing augmentation techniques~\citep{data_augmentation_survey}.
Early methods, including GraphCL~\citep{GraphCL} and InfoCL~\citep{InfoCL}, focus on random perturbations produced by dropping edges or nodes.
Other methods like graphMVP~\citep{graphMVP} and MICRO-Graph~\citep{MICRO-Graph} use domain knowledge to perform better than random augmentation.
More recent approaches such as AD-GCL~\citep{adgcl} and MEGA~\citep{mega} use a multilayer perceptron (MLP) as an augmentation learner.
They create augmented views by learning to drop low-relevance edges.
Similarly, RGCL~\citep{rational} adds a repulsive dynamic to the augmented views.

In parallel, considerable work has been done to interpret graph neural networks post training. A backpropagation of gradient-like signals~\citep{sensitivity}, perturbations of the input graph~\citep{gnnexplainer,PGexplainer}, and the training of a surrogate model~\citep{PGM-Explainer} are examples of such techniques.
These post-hoc techniques have been shown to be biased and unrepresentative of the true interpretations~\citep{stop_explaining}, leading to an increased interest in inherently interpretable models~\citep{GSAT,kerGNN}.
Furthermore, the aforementioned methods focus on interpreting predictions of models learned with supervision.
\citet{toward_exp_unsup_graph} have proposed USIB, the first post-hoc method to interpret unsupervised graph representations, but USIB faces the same faithfulness limitation as other post-hoc techniques.

Given existing limitations and the power of augmentation-based learning, we investigate how augmentation learning can faithfully and intrinsically provide interpretability.

\section{Materials and Methods}
\label{method}

In this section, we present INGENIOUS. We begin with a general overview and then provide a description of the augmentation and learning processes.

\subsection{Framework}
\label{subsec:architecture}

\begin{figure}[t]
    \centering
    \includegraphics[scale=0.20]{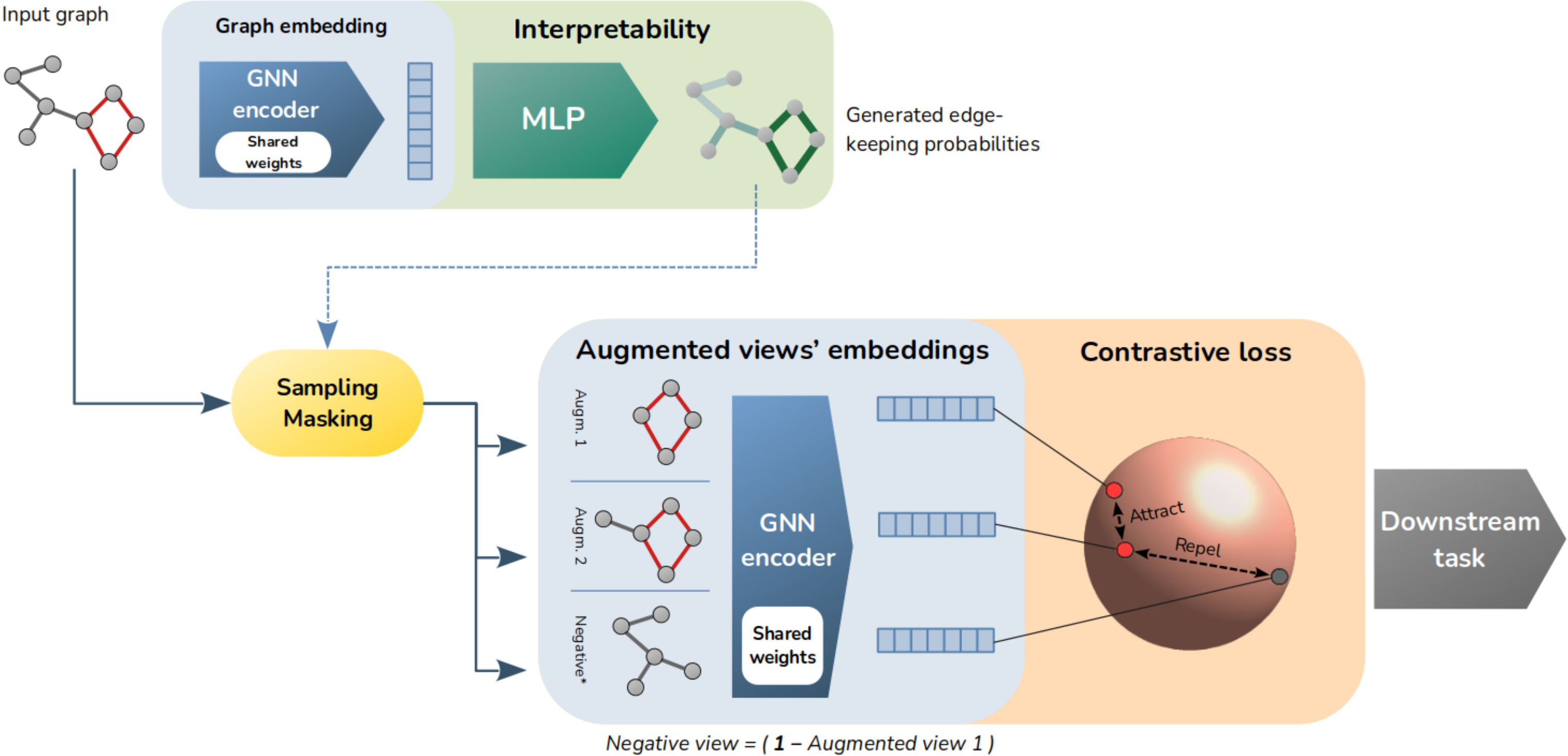}
    \caption{Schema of the INGENIOUS framework, as described in the text.}
    \label{fig:contrastive_training}
\end{figure}

We propose an inherently interpretable framework that produces graph representations in an unsupervised manner by leveraging learned graph augmentation, as shown in Figure~\ref{fig:contrastive_training}. Learned edge-dropping augmentation has been used both in contrastive learning by AD-GCL~\citep{adgcl} and MEGA~\citep{mega}, two recent state-of-the-art approaches, and in interpretability by GSAT~\citep{GSAT} and PGExplainer~\citep{PGexplainer}. Inspired by said contrastive approaches, INGENIOUS is trained with a contrastive loss, following a dual-branch approach as recommended by \citet{simclr}.
INGENIOUS is based on two modules: an embedding module that produces node embeddings ($\phi_u$) and graph embeddings ($\phi$), and an edge-selection module ($\theta$) that produces semantic-preserving sub-graphs by stochastically dropping uninformative edges. In this study, the embedding module is implemented as a graph neural network (GNN) encoder and the edge-selection module as an MLP, as described in Section~\ref{experimental_setup}.

The embedding module first generates an embedding $\phi_u$ for each node $u$ of the input graph. Using these node embeddings, the edge-selection module $\theta$ produces a keeping probability $p_{uv} \in [0, 1]$ for each edge $(u, v)$ of the graph. Then, the Gumbel-max reparameterisation trick is used to partially remove edges according to their keeping probability $p_{uv}$ in a differentiable manner. Each edge of an augmented view therefore has a weight $w_{uv} \in [0, 1]$. 

This edge-selection process is applied twice to obtain two positively augmented views.
A negatively augmented view is also produced by giving each edge the flipped weight $1-w_{uv}$ from the first positively augmented view.
The stochastic aspect of this data augmentation increases the robustness of the model, as it shows more diverse data to the model. Additionally, this aspect is needed to apply the dual-branch contrastive loss, since this loss necessitates at least two distinct augmented views for each graph.

Each augmented view of the input graph is then embedded using once again\footnote{When a module includes batch normalisation layers, it is tricky to use this same module for distinct steps in a model, because the input distribution of the batch normalisation layers may be different for each step the module is used for.
This aspect was not considered in previous approaches and may have perturbed their results and conclusions.
More details are given in supplementary materials.} the embedding module $\phi$.
Finally, a simclr loss is back-propagated to pull the embeddings of corresponding positively augmented views together while keeping the embeddings of other graphs' positively augmented views distant in the embedding space.
The negatively augmented view is also used as a part of the final loss, as described below.

At inference time, the Gumbel-max sampling is no longer used. Keeping probabilities $p_{uv}$ are therefore not used as probabilities anymore but are used directly as weights on edges of the original graph, i.e. $w_{uv} = p_{uv}$, in order for the augmentation to be deterministic. The full-graph embedding is defined as the embedding of the augmented view, to be used in any downstream task. This augmented view serves as the interpretation of the embedding.

\subsubsection{Embedding Module and Augmented-Graph Generation}
In this work, we consider attributed graphs $G = (V, E)$ where $V$ is the set of nodes and  $E \subset \{(u, v) | u \in V, v \in V\}$ is the set of edges of the graph. By giving an arbitrary order to $V$, each node can be represented by a single integer $u \in \mathbb{N}$. $G$ can then be described by its node-feature matrix $X \in \mathbb{R}^{|V|\times d}$ and its adjacency matrix $A \in \mathbb{R}^{|V|\times |V|}$, with $|V|$ the total number of nodes in the graph, $d$ the number of node features, $X_{ui}$ the value of the $i^{th}$ feature of the $u^{th}$ node $u$, and $A_{uv} = 1\text{ if }(u, v) \in E\text{ else } 0$. Node embeddings can be computed by the embedding module as $z_u =\phi_u\left(A, X\right)$. Graph embeddings can be computed as $z = \phi\left(A, X\right) = Pooling(\{z_u~\forall~u \in V\})$, with $Pooling$ an aggregation function on sets of vectors, such as the mean or the sum.

Augmented views of a graph $G$ can be produced in two different ways given the embedding $z_u$ of each of its nodes $u$. The first way, which was described briefly above, is this: The edge-selection module $\theta$ produces a keeping probability $p_{uv}=\theta([z_u, z_v])$ for each edge $(u, v)$, with $[\cdot, \cdot]$ the concatenation operator. Then, edges are sampled following the Gumbel-max reparameterization~\citep{gumbel}, giving each edge its weight
\begin{equation}
    w_{uv} = Sigmoid\left( (Logit(p_{uv}) + \epsilon_{uv}) /\tau \right)
\end{equation}
with $Logit$ the inverse function of the $Sigmoid$ function, $\epsilon_{uv} = \log(\alpha_{uv}) - \log(1 - \alpha_{uv})$ a random noise defined using $\alpha_{uv} \sim Uniform(0,1)$, and $\tau \in \mathbb{R}^+$ a temperature hyperparameter we set to $1$. The higher the value of $\tau$, the more uniform the weight distribution.

The second way is to produce a keeping probability $p_u = \theta(z_u)$ for each node, sample nodes by giving them a weight $w_u = Gumbel(p_u)$ with $Gumbel$ the Gumbel-max trick described above, and then lift the weights from nodes to edges following $w_{uv} = w_u \cdot w_v$.

Although both could be used for both graph and node embeddings, in our approach we use the first way for graph embeddings and the second way for node embeddings, since it works best empirically. These edge weights are then used to weigh messages during the message-passing operation of the GNN encoder.
At train time, independent samplings are used to generate two distinct positively augmented views.
The negatively augmented view is obtained by flipping the first positively augmented view, that is, each edge of the negatively augmented view has a weight $w_{uv} = 1 - w^+_{uv}$, with $w^+_{uv}$ the weight of the edge $(u, v)$ in the first positively augmented view.
Finally, the two positively augmented views and the negatively augmented view are embedded by the embedding module, which produces embeddings $z^+_1$, $z^+_2$ and $z^-$.
At inference time, the deterministic process described previously is used.

\subsubsection{Losses}

\label{methodo_losses}
In this section, we define the INGENIOUS loss. It is composed of four losses with diverse objectives.

The first loss is defined as the dual-branch approach of the simclr loss~\citep{simclr}, which leverages two augmented views rather than an anchor and an augmented view, as proposed by \citet{dual_branch}. This is the main loss of the framework as it structures the embedding space. Specifically, given augmented-view-embedding triplets coming from $N$ distinct graphs and represented as $\{Z^+_1, Z^+_2, Z^-\} \subset \mathbb{R}^{N\times D}$, with $D$ the embedding dimension, we first normalise each embedding so that it has an L2-norm of 1. Then, the loss is defined as
\begin{equation}
    \mathcal{L}_{simclr}(Z^+_1, Z^+_2) = -\frac{1}{2N}\sum_{i=1}^{N} \Big( \log(l_i(Z^+_1, Z^+_2)) + \log(l_i(Z^+_2, Z^+_1)) \Big)
\end{equation}
\begin{equation}
    l_i(Z^+_1, Z^+_2) = \frac{
        \exp(\operatorname{sim}(Z^+_{1i}, Z^+_{2i}) / \beta)
    }{
        \sum_{k=1}^{N} \Big(\exp(\operatorname{sim}(Z^+_{1i}, Z^+_{2k}) / \beta) + \mathbb{I}_{[k\neq i]} \exp(\operatorname{sim}(Z^+_{1i}, Z^+_{1k}) / \beta) \Big)
    }
\end{equation} 
with $\operatorname{sim}(\cdot, \cdot)$ a similarity operation such as the dot product, $\mathbb{I}_{[\text{condition}]}$ an indicator function with a value of 1 when the condition is met, else 0, and $\beta$ a hyperparameter we set to $0.07$ as in \citet{simclr}.

The second loss, which we call the $negative$ loss, complements the first loss in structuring the latent space. It leverages the negatively augmented views of the normalised triplets:
\begin{equation}
    \mathcal{L}_{-}(Z^+_1, Z^-) = \sum_{i=1}^N \mathcal{L}_{simclr}(Z^*_i, Z^*_i)
\end{equation}
with $Z^*_i = \{Z^+_{1i}, Z^-_i\} \in \mathbb{R}^{2\times D}$.
The motivation behind this idea stems from the observation that the simclr loss has an attracting effect between corresponding augmented-view embeddings and a repelling effect between everything else. If corresponding augmented-view embeddings are always an embedding and itself, there is no attracting effect anymore as something cannot attract itself. Thus, only the repelling effect of the simclr loss has an effect, which is a repelling effect between the positively-augmented-view embedding and the negatively-augmented-view embedding.

The third loss is the information loss from GSAT. It is used as regularisation. This loss minimises the mutual information between the positively augmented view and its original graph anchor following the graph information bottleneck (GIB) principle~\citep{GSAT}. It favours sparser augmented views as it prevents the model from collapsing and selecting all edges as important. The information loss is defined as:
\begin{equation}
\mathcal{L}_{info}(G, P) = \frac{1}{|E|}\sum_{(u, v) \in E} p_{u v} \log \frac{p_{u v}}{r}+\left(1-p_{u v}\right) \log \frac{1-p_{u v}}{1-r}
\end{equation}
with $P = \{p_{uv} \forall (u, v) \in E\}$ the keeping probabilities of all edges, $G = (V, E)$ a batch of graphs as one big graph, $|E|$ the total number of edges and $r \in [0, 1]$ a hyperparameter that represents the prior of the probability to keep any given edge.
Similarly to GSAT, the hyperparameter $r$ is initially set to 0.9 and gradually decays to the tuned value of 0.7. As proposed in GSAT, the weights $w_{uv}$ are used in practice instead of the probabilities $p_{uv}$.

The fourth and last loss is the watchman loss. \citet{augment_drop_info} have shown that edge-dropping augmentation can degrade the quality of the latent space. Additionally, we have observed that learned-augmentation-based methods sometimes have unstable training. Preliminary experiments supporting this claim are given in supplementary materials. The watchman loss mitigates these issues by forcing the embeddings to contain graph-level information. In detail, a watchman module $\psi$ uses augmented-view embeddings $Z^+_1$ to predict $\Lambda_k(G_i) \forall i$, the top $k$ eigenvalues of the Laplacian matrix of each graph $G_i$ composing the batch $G$. The loss is defined as
\begin{equation}
    \mathcal{L}_\text{Wm}(G, Z^+_1) = \frac{1}{N}\sum_{i=1}^{N}MSE(\psi(Z^+_{1i}),\Lambda_k(G_i))
\end{equation}
with $MSE(\cdot, \cdot)$ the squared euclidean distance between two vectors.
This loss is meaningful for graph-embedding tasks only, as node embeddings have no reason to contain information about the full graph.

Finally, the INGENIOUS loss is defined on a batch $G$ of $N$ graphs as: 
\begin{equation}
\mathcal{L} = \mathcal{L}_{simclr}(Z^+_1, Z^+_2) + \mathcal{L}_{-}(Z^+_1, Z^-) + \mathcal{L}_{info}(G, W) + \lambda_\text{Wm} \cdot \mathcal{L}_\text{Wm}(G, Z^+_1)
\end{equation}
with $W = \{w_{uv} \forall (u, v) \in E\}$ the edge weights of the first positively augmented view and $\lambda_\text{Wm}$ a hyperparameter we set to $0$ for node-embedding tasks and to $0.3$ for graph-embedding tasks, based on preliminary empirical results.

\section{Experiments}\label{experiments}
In this section, we evaluate INGENIOUS both in terms of utility and interpretability by comparing its performance to the performance of its unsupervised competitors AD-GCL and MEGA on graph-classification tasks and we conduct an ablation study on the loss terms. However, these competitors have not been extended to handle node-classification tasks, and to the best of our knowledge, there is no competitor that uses learned augmentation for these tasks. For both graph and node classification tasks, we use GSAT not as a direct competitor, as it is a supervised model, but as an upper bound on the obtainable utility.

\subsection{Experimental Setup}

\label{experimental_setup}
 
\paragraph{Datasets} For graph classification, we use two synthetic datasets (\textit{BA2Motifs}~\citep{PGexplainer}, \textit{SPMotifs.5}~\citep{spmotifs}) and one real-world dataset (\textit{Mutag}~\citep{TUDatasets}). For node classification, we use two synthetic datasets (\textit{Tree-grid}, \textit{Tree-cycle}~\citep{gnnexplainer}) and one real-world dataset (\textit{Cora}~\citep{Cora}). All of them are used with a batch size of 256 except for \textit{Cora} which is used with a batch size of 128. For \textit{Mutag} and \textit{BA2Motifs}, we randomly split the data into train (80\%), validation (10\%) and test (10\%) sets. For the other datasets, we use the same split as in their original papers.

\paragraph{Model} For the embedding module $\phi$, we use a GIN~\citep{GIN} model with 3 layers. For graph-classification tasks, a final pooling layer is added to the model. For the edge-selection module $\theta$, we use a 2-layer MLP. For the watchman module $\psi$, we use a 3-layer MLP. We use a learning rate of 1e-3 and a dropout ratio of 0.3, and we train the whole on 150 epochs. For all approaches, the metrics defined in Sections~\ref{utility} and \ref{subsec:interp} are measured on the model that obtains the best downstream ACC (see Section~\ref{utility}) measured on the validation set, amongst all epochs, as in GSAT. All results are averaged on 3 random seeds.

\subsection{Utility}\label{utility}
\begin{figure}[t]
    \centering
    \includegraphics[trim={0 0.2cm 0 0.22cm}, clip, scale=0.5]{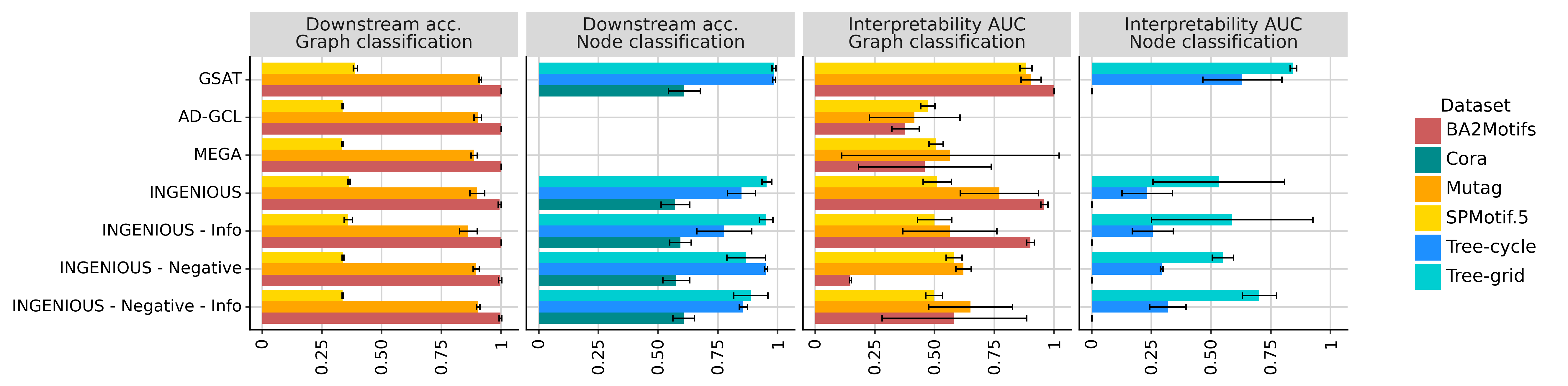}
    \caption{Downstream ACC and interpretability AUC. INGENIOUS outperforms its unsupervised competitors, AD-GCL and MEGA.}
    \label{aucs}
\end{figure}

The utility of embeddings learned without supervision is traditionally \citep{adgcl} approximated by the performance of simple models trained on these embeddings and their labels across various downstream tasks.
Similarly, we feed the representations to be evaluated to a linear model and we train the latter on the classification task associated with the available labels. The final metric is the accuracy of the model, which we call \textbf{downstream ACC}. As visible in Figure~\ref{aucs}, the accuracy of INGENIOUS is similar to the accuracy of its competitors (AD-GCL and MEGA) in the case of graph classification. Furthermore, it is close to the supervised upper bound (GSAT) despite being unsupervised.

\subsection{Interpretability} \label{subsec:interp}
 \begin{figure}[b]
    \centering
    \includegraphics[trim={0 1cm 0 1cm}, clip, scale=0.29]{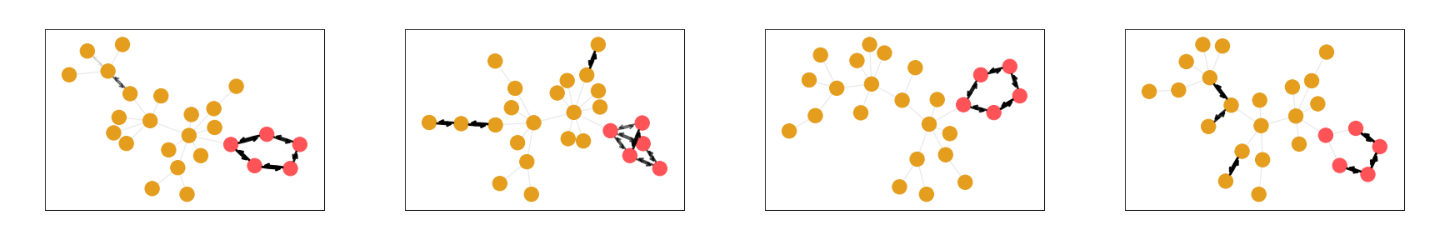}
    \caption{Illustrative examples of interpretations produced by INGENIOUS on four \textit{BA2Motifs} graphs selected randomly. Nodes included in the class-specific motifs are shown in red. Edge importance weights produced by INGENIOUS are represented through edge thickness. Therefore, thick edges between red nodes indicate that INGENIOUS focuses on the informative motifs.}
    \label{visu_interp}
\end{figure} 

The idea of interpreting representations can be defined as finding informative parts of the input that led to its representation.
Like most graph-interpretability methods \citep{taxo-survey}, INGENIOUS focuses on topology and produces sub-graphs as interpretations. Examples of such interpretations are visible in Figure~\ref{visu_interp}.
Interpretability and how to measure it has been extensively studied in the supervised setting \citep{interp_charac}, but little work has focused on interpretability for unsupervised representation learning. Monotony and localisation metrics have been introduced by \citep{RELAX,toward_exp_unsup_graph}, but most of the major categories introduced by the taxonomy proposed by \citet{interp_charac} have not been covered yet. Building on previous work, we introduce five metrics - \textit{Faithfulness}, \textit{Opposite faithfulness}, \textit{Wasserstein distance}, \textit{Bi-modality} and \textit{Interpretability AUC} -, all based on desirable characteristics for interpretations.

\paragraph{Correctness and Completeness} Correctness (how much the interpretation truly describes the behaviour of the model) and completeness (how much of the model behaviour is described by the interpretation)~\citep{interp_charac} are studied by evaluating whether an interpretation aligns with the edges that truly impact the embedding.
The canonical approach to evaluating it is the deletion (resp. insertion) experiment introduced by RISE~\citep{Rise}, which involves removing elements in order (resp. opposite order) of importance and studying the impact on model outputs.
Usually, this impact is measured as a difference in predicted scores. However, in the unsupervised setting, outputs are not predicted scores, but embeddings. Therefore, we propose to extend the metric by measuring this impact as the Euclidean distance between the embedding of the original graph and the embeddings of the perturbed graphs.

To evaluate correctness, we use an adapted faithfulness metric. Given a graph, we delete its edges in decreasing order of importance weights given by the edge-selection module. To achieve this, we group edges of similar importance into bins. In our experiments, we use 30 bins. Then, we successively remove bins of edges from the graph in decreasing order of importance, going from the full graph to a graph without any edge. The perturbed graphs are then embedded using the embedding module on its own, taking the importance weights $w_{uv}$ of the remaining edges into account during the message-passing operation of the module. The final metric is the area under the curve (AUC) of dissimilarity scores across different percentages of edges removed. For graph classification, the dissimilarity score is the Euclidean distance between the embeddings of the pruned graphs and the embedding of the initial graph. For node classification, we perform this experiment for each node, taking only the $k$-neighbourhood ($k$ being the depth of the model) of that node into consideration, and we take the average dissimilarity score. In order to ease comparisons between models, these dissimilarity scores are normalised by scaling, to be equal to 1 when all edges are removed. Removing important edges should change the embedding. Hence, removing them first should result in a larger AUC.

To evaluate completeness, we use the opposite faithfulness. It seeks to assess the complementary statement to faithfulness: Are the edges marked as unimportant truly unimportant? 
To measure it, we apply the same procedure as for the faithfulness metric, except edges are removed in increasing order of predicted importance.
In theory, removing unimportant edges should not impact the embedding. The AUC should therefore be small. 

To summarise the results intuitively, we focus on the faithfulness gap, that is, the difference in value between the faithfulness and the opposite faithfulness. A larger difference is better.
Results are presented in Table~\ref{table_res}. Faithfulness and opposite faithfulness results are presented in supplementary material.
The model we propose has a clear advantage, achieving a high faithfulness gap in all cases, even higher than the supervised GSAT model. These results demonstrate the superior correctness and completeness of INGENIOUS.

\begingroup\centering
\begin{table}[t]
\caption{Faithfulness and Wasserstein results. A higher faithfulness gap means the interpretations are more faithful, and a higher Wasserstein gap means the embedding space is more continuous in terms of interpretations.\label{table_res}}
\resizebox{\linewidth}{!}{%
\begin{tabular}{llllllll}
\toprule
                           Method & \multicolumn{3}{c}{Faithfulness Gap} & \multicolumn{3}{c}{Wasserstein Gap} \\
  &           BA2Motifs &            Mutag &      SPMotifs.5 &                  BA2Motifs &            Mutag &      SPMotifs.5 \\
\midrule
        GSAT &  $.34 \pm .00$ &  $.21 \pm .08$ &   $.06 \pm .02$ &  $.04 \pm .00$ &  $\mathbf{.02 \pm .00}$ &  \uline{$.04 \pm .00$} \\
        AD-GCL&	$.01 \pm .13$	&$.03 \pm .09$	&$.08 \pm .06$	& $\mathbf{.10 \pm .01}$	&$.01 \pm .00$	&$.03 \pm .00$ \\
         MEGA&	$.06 \pm .02$	&$.39 \pm .12$	& $\mathbf{.68 \pm .03}$	&$.04 \pm .01$	&$.01 \pm .00$	&$.01 \pm .00$ \\
        
          INGENIOUS &  $.60 \pm .02$ &  \uline{$.41 \pm .18$} &   $.45 \pm .03$ &  \uline{$.06 \pm .01$} &  $.01 \pm .00$ &  $\mathbf{.05 \pm .01}$ \\
        INGENIOUS - Info &  $\mathbf{.68\pm .02}$ &  $\mathbf{.48 \pm .03}$ &   \uline{$.48 \pm .06$} &  $.05 \pm .02$ &  $.01 \pm .00$ &  $\mathbf{.05 \pm .01}$ \\
        INGENIOUS - Negative &  $.13 \pm .02$ &  $.18 \pm .00$ &   $.22 \pm .06$ &  $.02 \pm .01$ &  $.01 \pm .00$ &  $.02 \pm .01$ \\
         INGENIOUS - Negative - Info &  \uline{$.61 \pm .10$} &  $.15 \pm .09$ &   $.32 \pm .00$ &  $.03 \pm .01$ &  $.01 \pm .00$ &  $.03 \pm .00$ \\

\bottomrule
\end{tabular} 
} 
\end{table}
\endgroup

\paragraph{Continuity}
Continuity refers to the smoothness of the interpretability function~\citep{interp_charac}. We seek to evaluate whether inputs for which the model response is similar, that is, inputs with similar embeddings, also have similar interpretations.
However, much like faithfulness, no metrics have been defined in the literature to evaluate this in an unsupervised setting.
To evaluate it anyway, we need to define \textit{(dis)similarity}, both between embeddings and between interpretations. While embeddings lie in a Euclidean space, where the L2-norm can be used as a dissimilarity measure, the interpretations lie in the space of graphs. There is no consensual method to compare any two graphs or sub-graphs. Indeed, a large variety of methods have been proposed~\citep{netrd}. Amongst them, we choose to use the node-degree-based Wasserstein distance as a dissimilarity measure, as it is amongst the simplest to compute and the most intuitive.
In line with the literature on graph interpretability and augmentation, which focuses solely on graph topology, this measure disregards node features. It also has the advantage of enabling its use with any graph dataset.
The continuity metric we propose could be adapted with any other similarity metric with minor changes, but the adaptation would not significantly alter its theoretical analysis. The study of possible adaptations is therefore left for future work.

Formally, given a graph $G$, we obtain its interpretation $G^{exp}$ by applying a hard-threshold.
We keep a fixed percentage of the most important edges, arbitrarily set to 10\%. However, for datasets for which there is a prior on the number of important edges for a downstream task, we use that prior number instead of 10\%.

The Wasserstein distance (WD) between two graphs $G_1$ and $G_2$ serves as the dissimilarity between their interpretations and is defined as $d_W(G_1, G_2) = W_1(\nu(G^{exp}_1), \nu(G^{exp}_2))$, with $\nu(G)$ the distribution of node degrees in $G$ and $W_1$ the 1-Wasserstein metric. 
Let $\pi(G)$ be the local neighbourhood of a graph $G$, defined as the 10\% closest graphs of the dataset in the embedding space according to the Euclidean distance. To evaluate the continuity of interpretations, we compare two values: 

\begin{itemize}
    \item $W_1^{local} = \mathbb{E}_{i} \left[ \mathbb{E}_{G_j \in \pi(G_i)} \left[d_W(G_i,G_j)\right]\right]$, the expected WD between a graph and another graph in the local neighbourhood of the first one;
    \item $W_1^{global} = \mathbb{E}_{i} \left[ \mathbb{E}_{j} \left[ d_W(G_i,G_j)\right]\right]$, the expected WD between any two graphs.
\end{itemize}

To save time, the expected values are computed on a subset of 6 batches of graphs, i.e. 1536 graphs. A big difference between these two values would be a good indicator of continuity, as it would mean that graphs that are similar in the embedding space have interpretations that are more similar to each other than to the interpretations of other graphs.
To summarise the results intuitively, we focus on the difference $W_1^{global} - W_1^{local}$; a larger gap indicates a stronger continuity. We report these differences in Table~\ref{table_res}. We see the advantage provided by the negative term of the INGENIOUS loss to produce an embedding space with continuous interpretations, as visible by the higher scores of INGENIOUS and INGENIOUS - Info compared to INGENIOUS - Negative  and INGENIOUS - Negative - Info, on \textit{BA2Motifs} and \textit{SPMotifs.5}. As opposed to \textit{Mutag}, these datasets have clear distinct motifs by class and therefore most benefit from the repulsive effect of the negative loss.

\paragraph{Compactness and readability}
Previous work has shown that sparsity is an important factor in human understanding \citep{interp_charac}, as it is hindered by human cognitive capacity limitations, and that interpretations should be sparse to not overwhelm the user. 
Moreover, for inherently interpretable methods as INGENIOUS, sparsity is related to faithfulness as non-sparse interpretations would mean that all features are used by the model to produce the embeddings.
Nevertheless, we argue that sparsity, alone, can be misleading, since it can favour interpretations with globally lower weights.
An ideal interpretation should be bi-modal, with weights of 1 for important edges and 0 for unimportant ones. We, however, relax this assumption and consider that an interpretation is easy to read if the distribution of weights has two modes. To estimate this, we approximate the density function of edge weights by using a histogram with ten bins between 0 and 1. We also define the \textbf{sparsity index} as the average of the edge importance weights. 

\begin{figure}[t]
    \centering
    \includegraphics[trim={0 0.22cm 0 0.21cm}, clip, scale=0.46]{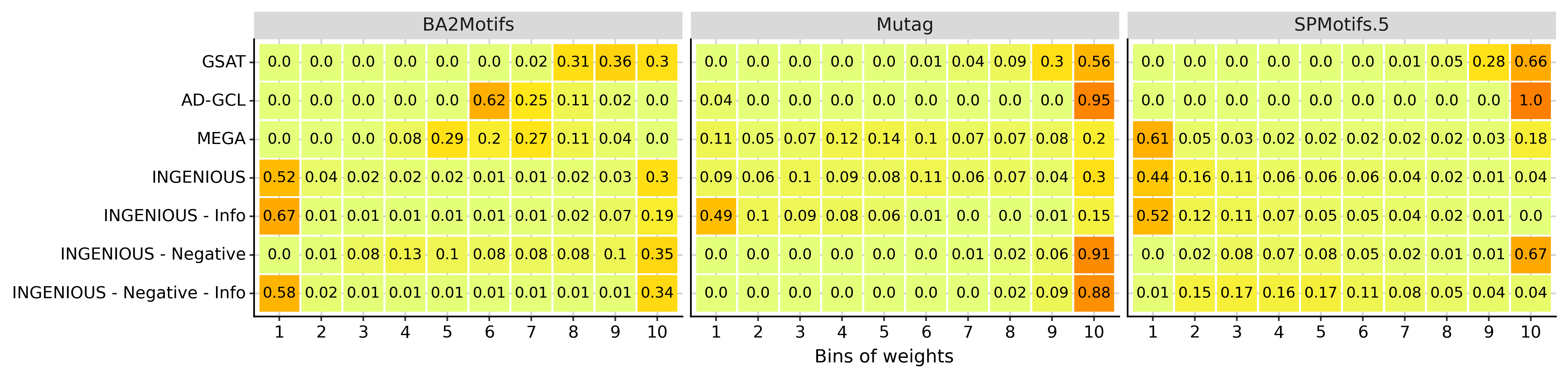}
    \vspace{-2em}
    \caption{Density function of edge weights. Optimal interpretations should have two density peaks and maximal separation. INGENIOUS results in importance values that are much sparser and closer to extreme values.}
    \vspace{-1em}
    \label{sparsity_heatmap}
\end{figure}

Figure~\ref{sparsity_heatmap} shows that on simple datasets such as \textit{BA2Motifs}, all methods produce sparse and bimodal interpretations, even without regularisation. However, more complex datasets such as \textit{Mutag} and \textit{SPMotifs.5} highlight the need for regularisation. On \textit{Mutag} for example, adding the regularisation lowers the proportion of scores between 0.9 and 1 from 90\% to 30\%. Both AD-GCL and GSAT tend to collapse on high value and narrow-support densities, as opposed to MEGA which has importance scores closer to extreme values.

\paragraph{Coherence}
Coherence assesses the alignment with domain knowledge.
Sometimes, in the supervised setting, human prior can be used to define expected interpretations that can be compared with generated interpretations. In the literature, this is referred to as \textit{the clicking game} \citep{RELAX}.
However, it relies on the prior that the model focuses on the same features as the domain expert would, which is a strong assumption~\citep{Rise}.
In an unsupervised setting, it is even harder to define relevant expected interpretations for embeddings. 
A way to approach this is to use synthetic datasets only; we use \textit{BA2Motifs}, \textit{SPMotifs.5}, \textit{Tree-grid} and \textit{Tree-cycle}. 
In these datasets, base graphs are generated from the same distribution. Then, different motifs are added to each random graph \citep{spmotifs,gnnexplainer,PGexplainer} depending on the dataset. 
For graph classification datasets, the motif also depends on the label of the graph. Therefore, we hypothesise that the base graph can be considered as noise and that all the distinctive information of each graph is in the motifs.

Under this hypothesis, any good unsupervised embedding method should rely on the motifs only. This is partially validated by the good accuracy shown in Section~\ref{utility}, as it is impossible to have good accuracy on the synthetic datasets if the embeddings contain no information about the motifs. The motifs can therefore be used as expected interpretations.

To measure how close a produced interpretation is to the expected one, we consider an edge-classification problem. An edge is labelled as important if it is part of the expected interpretation. The predicted probability of an edge being important is given by the produced importance weight of that edge. The area under the ROC curve is reported in Figure~\ref{aucs}. In this paper, we call this metric the \textbf{interpretability AUC}.

In terms of interpretability AUC, INGENIOUS outperforms its unsupervised competitors. The supervised approach GSAT is better as expected.
The ablation study shows that both regularisation losses are needed to reach a better interpretability AUC, although their advantage is less significant on node classification tasks. 
This could find an explanation in the link between interpretability and sparsity. We study this link in Section~\ref{sparsity_analysis}.

\subsection{In-Depth Analysis of the Sparsity}
\label{sparsity_analysis}

\begin{figure}[h]
    \centering
    \includegraphics[trim={0 0.3cm 0 0.22cm}, clip, scale=0.3]{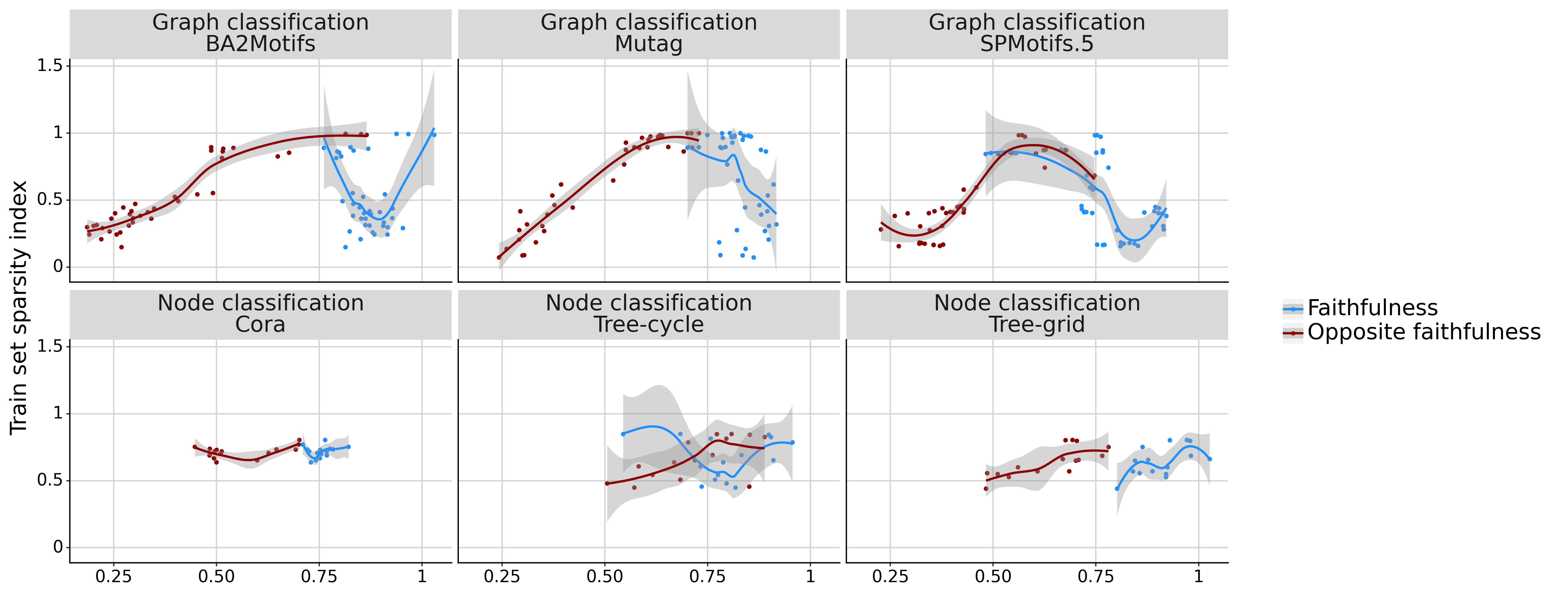}
    \vspace{-0.5em}
    \caption{(Opposite) faithfulness with respect to the sparsity index. A lower sparsity results in a lower opposite faithfulness (lower is better) and a marginally higher faithfulness, which means a better faithfulness gap. The grey area represents the 95\% confidence interval for a smoothed conditional mean.}
    \vspace{-1em}
    \label{faithfulness_sparsity}
\end{figure}

\paragraph{Impact of the sparsity on faithfulness} In addition to being a desirable characteristic for interpretations, sparsity impacts contrastive learning by increasing the difference between the augmented views.
Figure~\ref{faithfulness_sparsity} shows a significant positive correlation between opposite faithfulness (red dots) and sparsity, while there is a negative correlation between faithfulness (blue dots) and sparsity. This suggests there may be a link between higher sparsity and better faithfulness gap. This conclusion applies to all graph-classification tasks. 
For node-classification tasks, the sparsity has a smaller range of values and presents no correlation with the faithfulness but still a small correlation with the opposite faithfulness.
Further results with a random baseline are presented in supplementary materials.
Overall, the faithfulness gap is improved by the sparsity.

\paragraph{Impact of hyperparameters on sparsity} This section deals with the impact of the temperature of the Gumbel-max sampling and the impact of the edge-keeping-probability prior $r$.
A high temperature drives the sampling probability closer to a uniform distribution, which increases the diversity of augmented views. Figure~\ref{temperature} shows that sparsity decreases with an increasing temperature value until it reaches a plateau. On the synthetic dataset \textit{BA2Motifs}, a temperature sweet-spot for interpretability AUC is visible at a temperature equal to 20; too high or too low of a temperature is harmful. However, for the other experiments of this paper, to ensure the comparability of our results with those of GSAT, we use the same temperature value as in GSAT, i.e. $\tau = 1$, even though this choice might not result in optimal performances.

\begin{figure}[t]
    \centering
    \includegraphics[trim={0 0.2cm 0 0.16cm}, clip, scale=0.29]{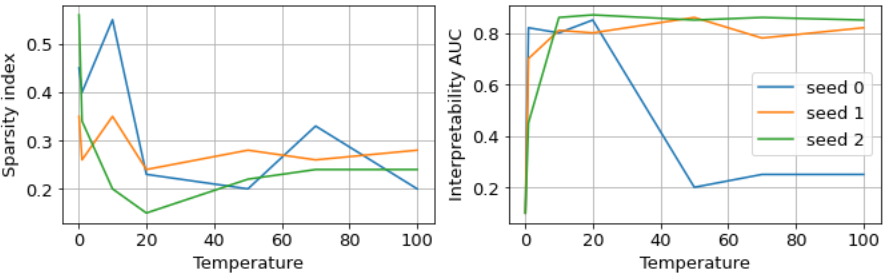}
    \vspace{-0.5em}
    \caption{Sparsity and interpretability AUC on \textit{BA2Motifs} with different temperatures. A sweet spot is visible at 20; too high or too low temperatures worsen the results.}
    \vspace{-1em}
    \label{temperature}
\end{figure} 

\begin{figure}[h]
    \centering
    \includegraphics[trim={0 0.22cm 0 0.22cm}, clip, scale=0.45]{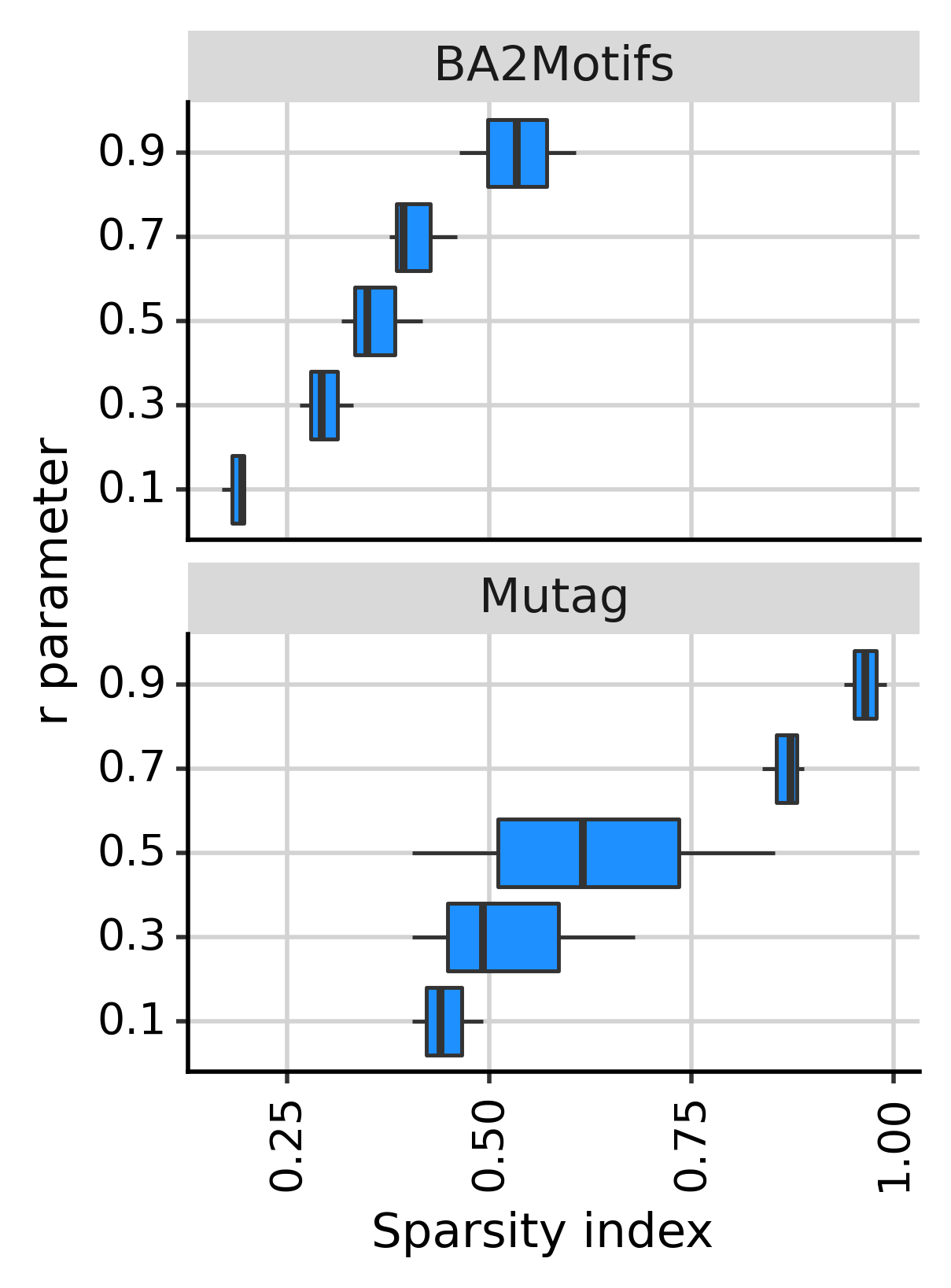}
    \includegraphics[trim={0 0.22cm 0 0.22cm}, clip, scale=0.45]{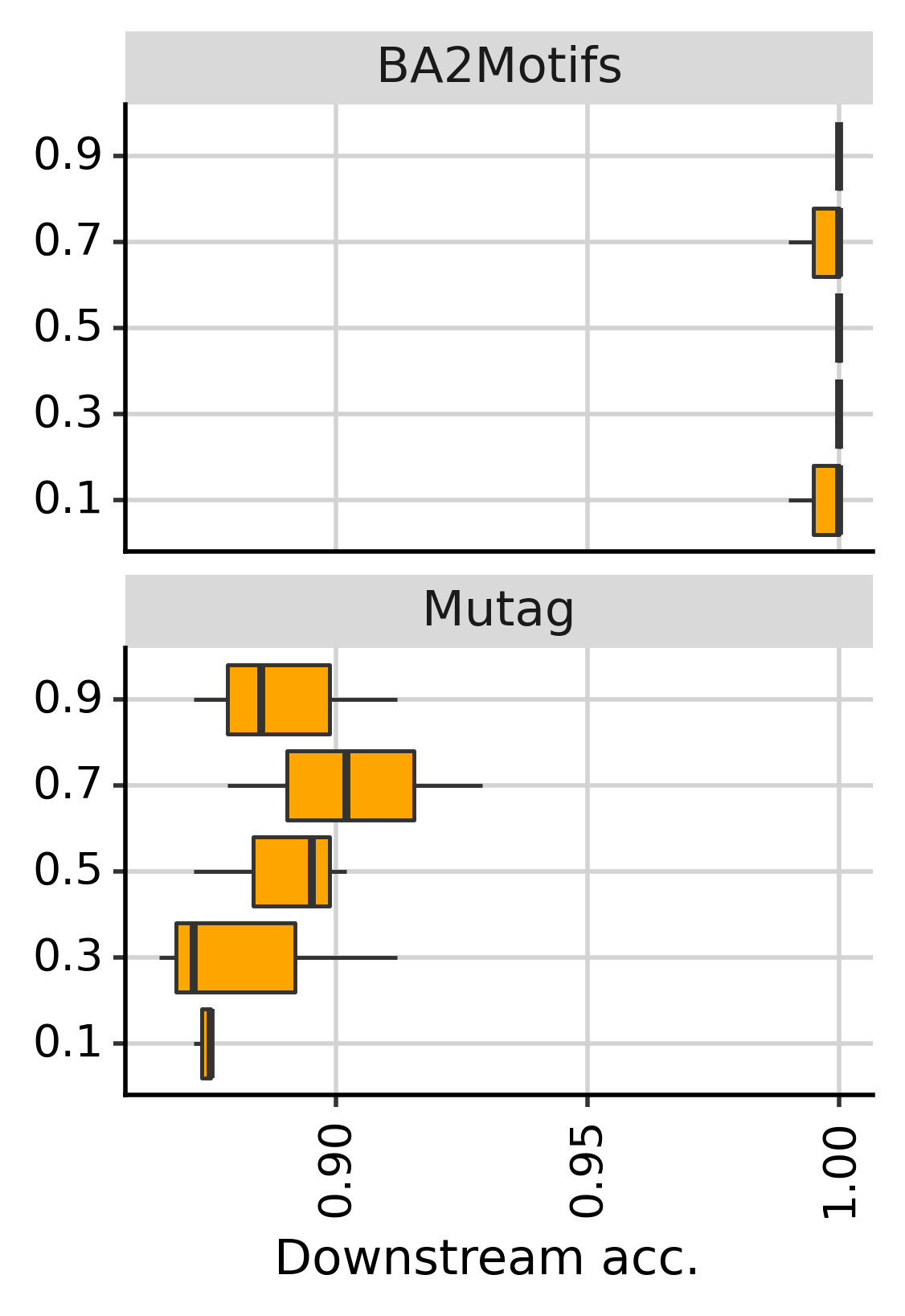}
    \includegraphics[trim={0 0.22cm 0 0.22cm}, clip, scale=0.45]{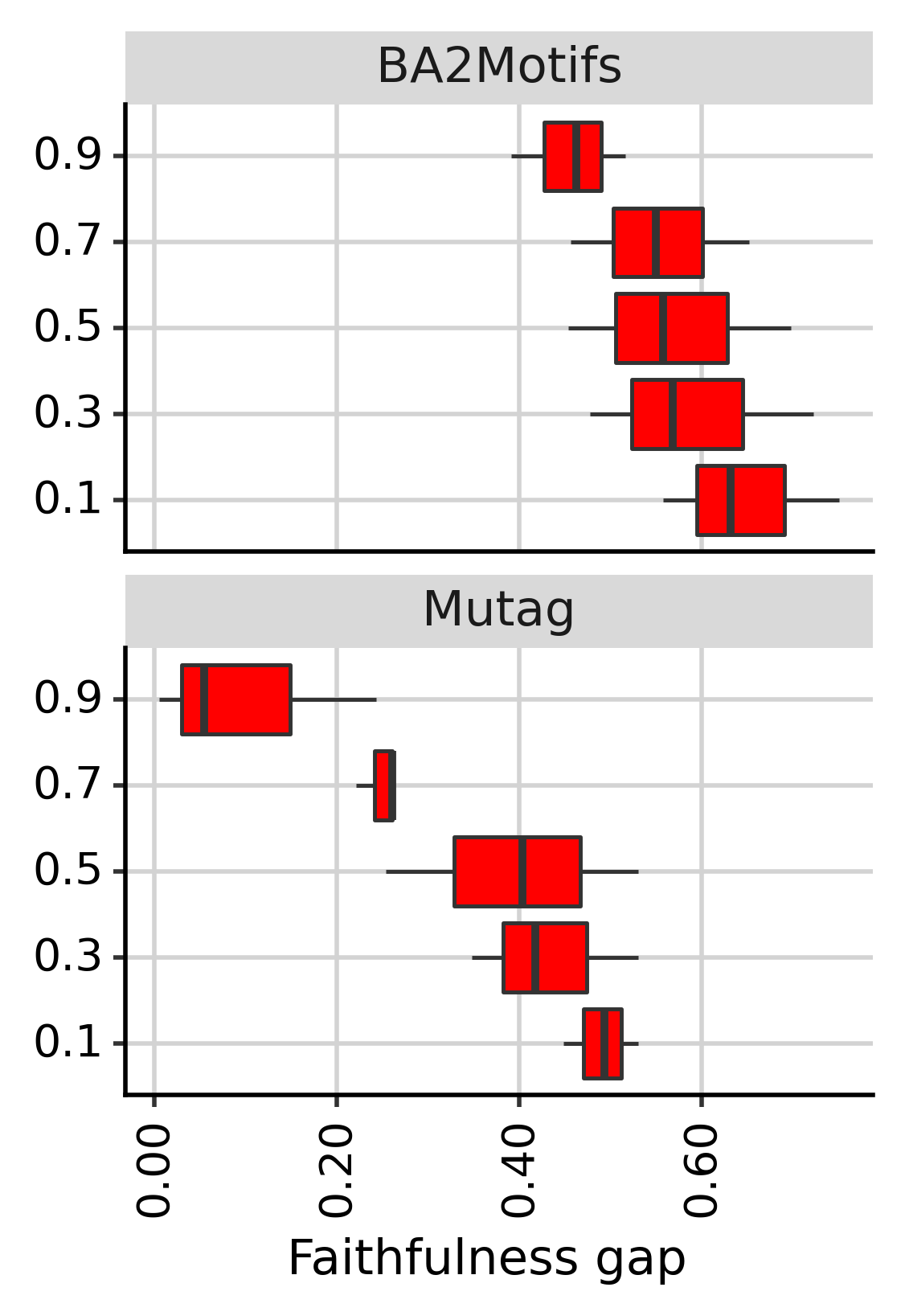}
    \includegraphics[trim={0 0.22cm 0 0.22cm}, clip, scale=0.45]{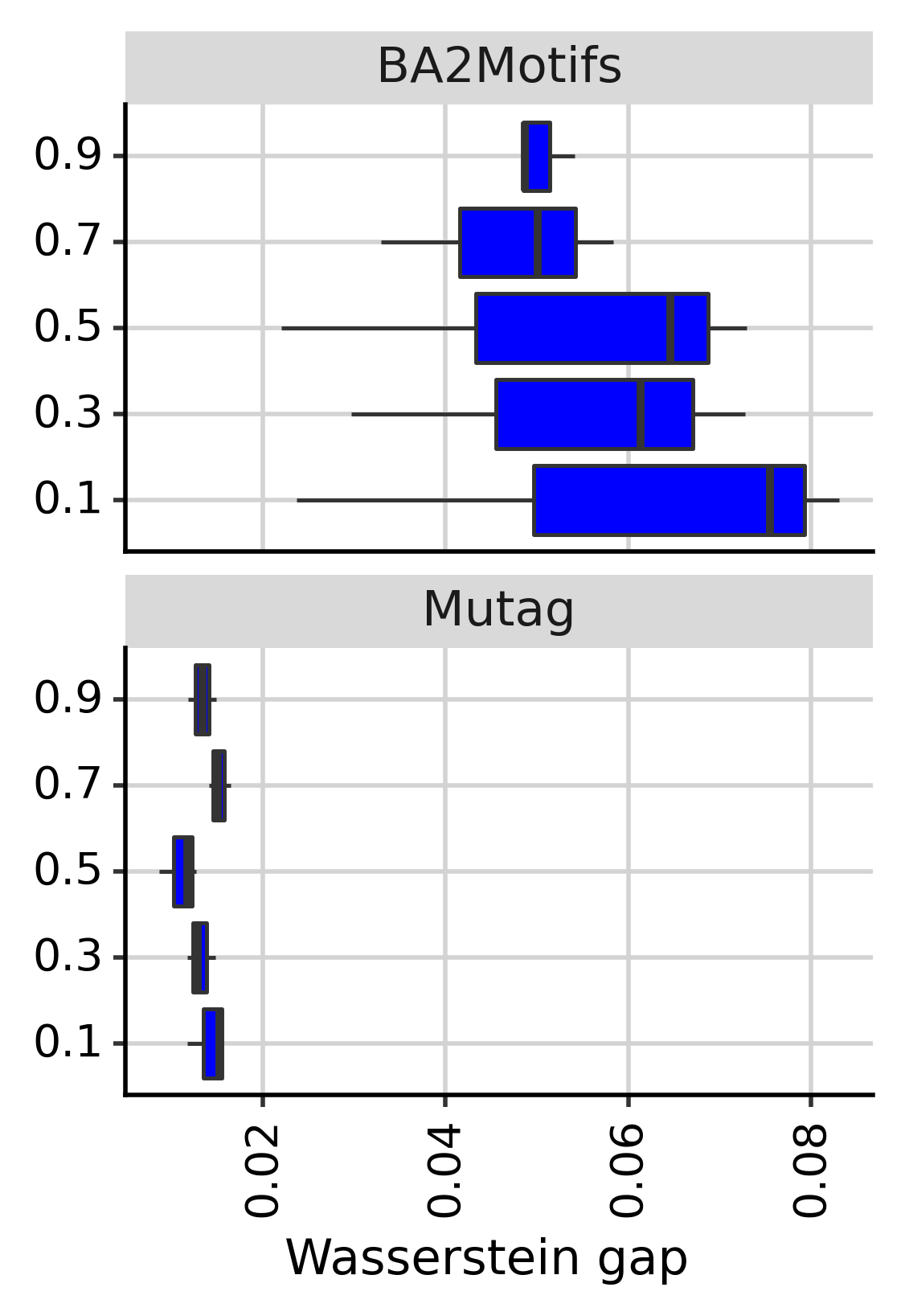}
    \vspace{-0.5em}
    \caption{Analysis of the hyperparameter $r$.
    A lower $r$ hyperparameter can improve the sparsity, resulting in interpretations that are more correct, complete and robust, with only a slight impact on the downstream ACC.
    }
    \vspace{-1em}
    \label{explore_r}
\end{figure}

The first column of Figure~\ref{explore_r} shows that $r$ is positively correlated with the sparsity of INGENIOUS. Lower $r$  leads to a better faithfulness gap, meaning more complete and correct interpretations.
Sparser augmented views lead the model to use fewer edges to produce embeddings, which results in a slight drop in downstream ACC. This phenomenon is less marked on simple datasets where all methods reach a perfect downstream ACC. This trade-off between downstream ACC and interpretability has already been documented in previous work~\citep{trade_off}.
Similarly, the sparsity positively impacts the continuity of the interpretability function, as can be seen by the increasing Wasserstein gap. This impact, however, is less strong on \textit{Mutag}.
In INGENIOUS, the sparsity can be controlled to an extent through regularisation and hyperparameter tuning. 

\section{Limitations and Future Work} 

Like most GNN-interpretability methods \citep{PGexplainer,gnnexplainer,taxo-survey}, we limit our interpretability analysis to topology. However, early results in supplementary materials show the possibility of extending INGENIOUS to node features.
As defined by \citet{KIM}, we limit our study to a functionally grounded evaluation (which does not require human subjects). Both human-grounded and application-grounded evaluations, which are more specific and costly, are left as future work.

\section{Conclusion}

We unify previous works on augmentation-based graph representation learning to introduce a new unsupervised embedding approach. Our method, INGENIOUS, introduces new losses for better regularisation. We show their positive impact through an ablation study.
Aiming for interpretability, we introduce new metrics to evaluate the quality of interpretations of graph representations learned without supervision, building upon previous work in the supervised setting.
Thanks to the approach and metrics we propose, we show that it is possible to produce inherently interpretable embeddings that are also useful for various downstream tasks and that augmented views, when sparse, can be used as interpretations.
INGENIOUS compares favourably with state-of-the-art unsupervised models in terms of utility of the embeddings and correctness, completeness, continuity and readability of the interpretations.
By tuning a reduced set of two hyperparameters, the user can manage the sparsity/utility trade-off. 
Our work paves the way for novel approaches of intrinsically interpretable unsupervised embeddings for modalities as complex as graph data.

\newpage
\bibliography{refs}
\newpage
\appendix

\section{Reusing a Module with Batch Normalisation}
\label{app:batchnorm}
All our competitors use approximately the same GIN architecture for their experimentation~\citep{adgcl,mega,GSAT}.
Specifically, they all use batch normalisation layers, and so do we.
However, considering the context it is used in, we argue that the way it should be used to be mathematically correct is not as straightforward as the way it is used in these approaches.
This may have unexpected effects on their results and conclusions.
Therefore, we propose a batch-norm switch to fix the issue, and we use it for our approach.
As this is not within the scope of this paper, only a preliminary experiment is presented to demonstrate the relevance of a switch in practice.
Further investigation into alternative methods to address the problem and the resulting effects on performances is left for future work. 

\subsection{Theoretical Analysis}
In more detail, all mentioned approaches have one point in common: they use the same neural network to embed both the original graph ($raw$) and an augmented view of it ($aug$).
However, the original graph and its augmented view come from different distributions, with significant differences.
For example, the augmented view usually has lower node degrees, as only a subset of the original graph edges is kept.
Ultimately, this difference has an impact on the input of the batch normalisation layers of this neural network.

As a reminder, the batch normalisation layer uses the batch mean and variance to normalise its input,
and stores a running mean of these values to be used at test time as an estimate of said values.
That is, the normalisation performed at train time is meant to be roughly equivalent to the normalisation performed at inference time.
In that way, given an input, it produces roughly the same output, whether it is in train or in inference mode.
This is important as the end of the network is trained on that output.

In our case, the mean and variance of the input of the layer are different whether the input of the network is a raw graph $(\mu_{raw}, \sigma_{raw})$ or an augmented view of it $(\mu_{aug}, \sigma_{aug})$.
At train time, the normalisation will therefore be different in both cases.
However, at our competitors' inference time, the estimate to perform the normalisation is something close to the average of the real values $(\frac{1}{2}(\mu_{raw} + \mu_{aug}), \frac{1}{2}(\sigma_{raw} + \sigma_{aug}))$, for both the original graph and its augmented view.
As a result, if the means and variances are very different, the outputs will not be similar in any way to the ones at train time.
The final output of the network will therefore be highly unpredictable.

\subsection{Batch-Norm Switch and Preliminary Experiment}
A way to avoid this issue is to have two different running means: one for the original graph distribution, and one for the augmented graph distribution.
We call this solution a batch-norm switch, as it switches between running means.
With a batch-norm switch, the training of the model remains the same as without the switch, but the inference is mathematically correct.
To assess the relevance of the switch, a quick experiment is presented hereunder, based on the code provided by GSAT authors for the \textit{BA2Motifs} dataset with a $GIN$ architecture, using seed $0$ only.

In GSAT, the issue is mitigated by having augmented views close to the original graphs, i.e. removing on average less than half the edges.
It is further mitigated by an early stopping criterion based on a metric on the validation set, which is unlikely to give good results if the network produces unpredictable results.
The early stopping will therefore tend to select epochs for which the issue has no negative impact.

\begin{figure}[h]
\begin{center}
\subfigure[Interpretability AUC]{
    \includegraphics[width=0.45\textwidth]{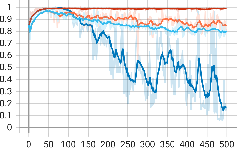}
}
\subfigure[Downstream ACC]{
    \includegraphics[width=0.45\textwidth]{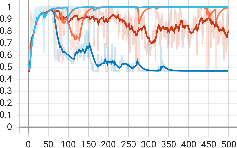}
}
\subfigure[Avg. Background Attention Weights]{
    \includegraphics[width=0.45\textwidth]{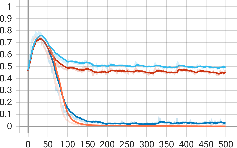}
}
\subfigure[Avg. Signal Attention Weights]{
    \includegraphics[width=0.45\textwidth]{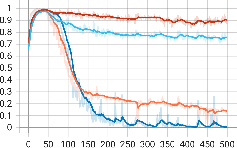}
}
\caption{Comparison of GSAT with a batch-norm switch (orange and teal) and without a batch-norm switch (red and dark blue) on \textit{BA2Motifs}.
The final value of $r$ is set to $0.1$ (dark blue and orange) or $0.5$ (teal and red).
A smoothing effect is applied to ease the analysis.
Non-smoothed results are shown in transparent.}
\label{fig:batchnormswith}
\end{center}
\end{figure}

For our analysis, we change these two parameters.
First, we run the experiments for $500$ epochs instead of $100$ and show the full curves observed during training rather than the final result of the selected epoch.
Second, for half the experiments, we use a final $r$ value of $0.1$ instead of $0.5$.
This means that starting from epoch $80$, roughly $10\%$ of edges are kept rather than $50\%$.
Note the first $40$ epochs are not affected by this second change, as GSAT adopt a step decay for $r$.
The result is shown in Figure~\ref{fig:batchnormswith}, measured on the test set.

As can be seen, the downstream ACC is greatly improved by the switch.
On the other hand, the interpretability AUC is improved by the switch when interpretations are sparse, but it is not as good when interpretations are not sparse.
We think it is because the unexpected effect aligns with some dataset properties.
For example, it could favour highly connected interpretations, which is good in the case of the \textit{BA2Motifs} dataset but may not be for other datasets.

\begin{figure}[!h]
\begin{center}
\includegraphics[width=\textwidth]{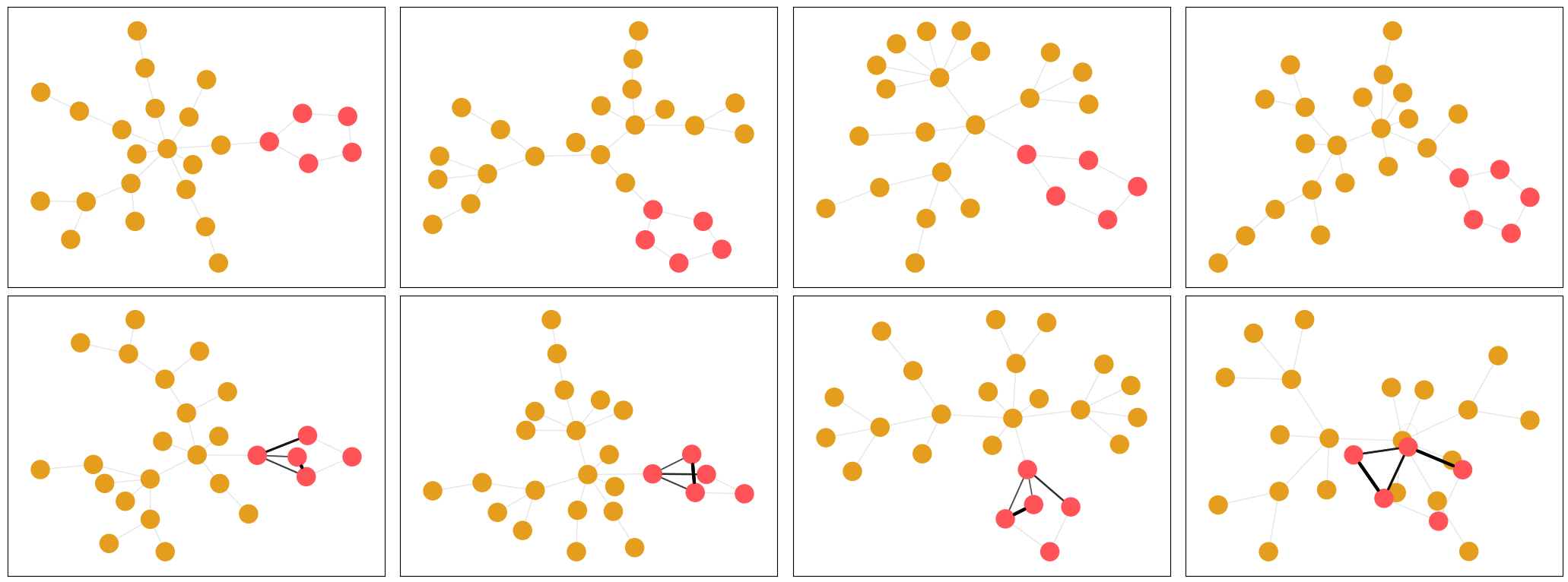}
\caption{Interpretations of GSAT with a batch-norm switch and final $r$ at $0.1$ at epoch 370 for class 0 (top) and class 1 (bottom). Red nodes are part of the motif, and bold lines are predicted as important.}
\label{fig:gsat_expl}
\end{center}
\end{figure}

Additionally, we can see that the model with a switch and sparse interpretations (the orange curves) has on average almost null attention weights on non-important edges ($0.0016$ on average at the end), but significant weights on important edges ($0.12$ on average at the end). 
This indicates that the model misses important edges, but never includes non-important edges.
For example, it could select length-3 cycles from class 1 and nothing from class 0, as shown in Figure~\ref{fig:gsat_expl}.
This could be considered a perfect interpretation, but it obtains a not-so-good interpretability AUC, given the expected interpretations include the length-5 cycles for each class.

A deeper analysis could be done to verify our hypotheses, challenge the batch-norm switch against alternatives, analyse results obtained with a batch-norm switch on each of our competitors or even challenge the \textit{BA2Motifs} dataset itself, but this is out of the scope of this paper and is therefore left for future work.

\newpage
\newpage

\section{Beyond Topology-Based Interpretations}
Our results indicate that there is a link between contrastive graph augmentation and interpretability. In such approaches, however, augmentations only focus on edges, and thus interpretations only highlight topology.
A natural question is whether these conclusions stand for node features too. This section presents some early work on the matter.

\begin{figure}
    \centering
    \includegraphics[scale=0.5]{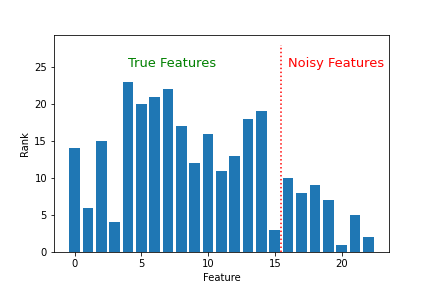}
    \caption{Ranking according to the average importance score given by the feature-augmentation module to each feature. It shows that noisy features are given less importance on average.}
    \label{features_selection}
\end{figure} 

We enhance the framework with an additional 2-layer MLP $\nu$, trained to learn importance weights for node features.
To make these weights understandable, we apply the feature augmentations in the input space, before being processed by the encoder.

The augmented feature matrix ($\widetilde{X}$) is obtained as $\widetilde{X} = (\nu(X) + \epsilon) \cdot X$ and with $\epsilon$ sampled as indicated in Section $3.1$ with  $\nu(X)$ a weight per node per feature, i.e. $\nu(X) \in \mathbb{R}^{N \times d}$.

We add an information loss on the importance scores to regularise the optimization, using the same $r$ as the one for the topology information loss.
As no ground truth is available for feature importance, we artificially add 20\% of noisy features. We then average the importance score obtained by each feature on the test set and rank them.
On \textit{Mutag}, with our complete loss, we obtain the following results: 

Figure~\ref{features_selection} shows that the noisy features are given less importance on average, thus demonstrating an intrinsic feature selection ability which can be used as a feature importance scheme.

\section{Impact of Sparsity on Human Readability}

\begin{figure}[h]
\begin{center}
\subfigure[Sparse importance weights]{
    \label{subfig:continuous}
    \includegraphics[width=0.45\textwidth]{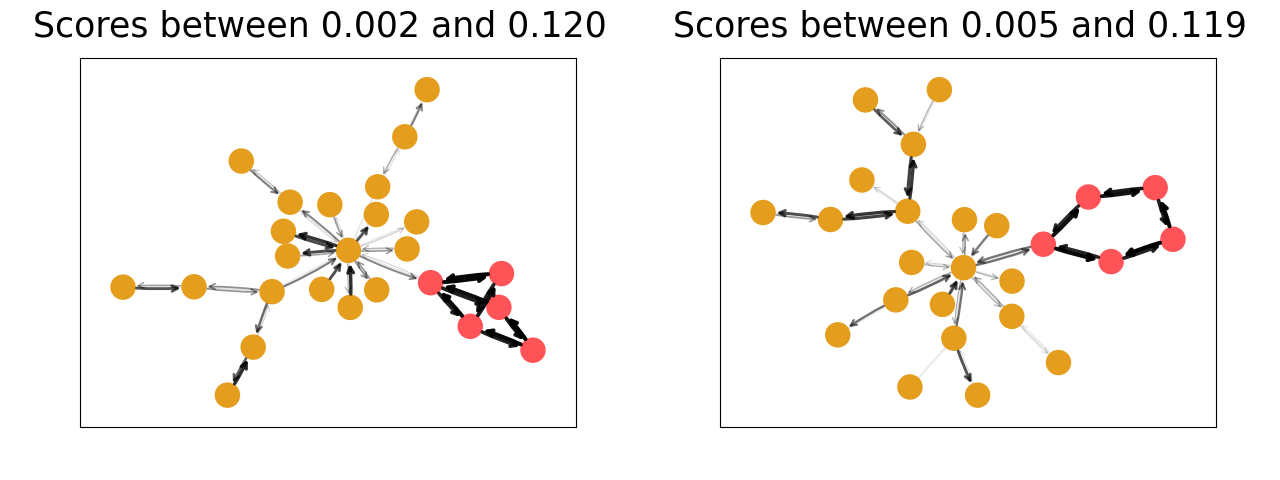}
}
\subfigure[Bi-modal importance weights]{
    \label{subfig:bimodal}
    \includegraphics[width=0.45\textwidth]{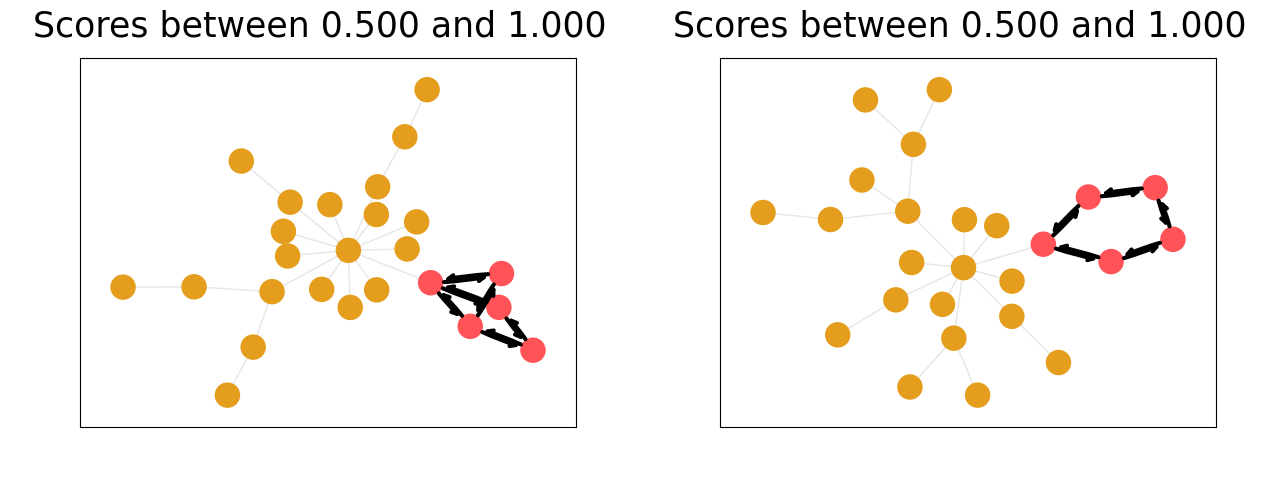}
}
\caption{Examples of importance weights. A min-max normalisation is applied for clarity. Original weight limits are shown as the title of the figures.}
\label{fig:bimodality}
\end{center}
\end{figure}

Figure \ref{fig:bimodality}(a) show some interpretations where important scores and unimportant ones are close. While truly important edges are highlighted, it is harder to distinguish them than in Figure \ref{fig:bimodality}(b)

\newpage
\section{Ablation Study of the Watchman Module}
Our framework introduces a watchman module, with the purpose to stabilize the optimization. We observe in Table \ref{downstream_graph}  that the presence of this module prevents quick drops in both interpretability AUC and downstream ACC, except for a few rare cases (as visible in Table~\ref{downstream_graph} and Table~\ref{table_res}).

This pattern is either positive or not significantly different across datasets, except for a few:  INGENIOUS - Negative on \textit{BA2Motifs} and certain methods on \textit{SPMotifs.5} which significantly decrease due to conflicting optimization objectives, however we advise to always use the watchman for our regularised loss INGENIOUS ($\mathcal{L}$) as it improves the results on all datasets.

Interestingly, the loss which benefits the most from the watchman is the simple unregularised simclr loss ($\mathcal{L}$-info-negative), probably because in this case, the problem is under-constrained. It acts as a regularizer by enforcing the recovery of eigenvalues from the embeddings. This result shows the importance of regularisations of any kind for contrastive learning.

We conclude that while performance diminishes with overfitting as we just discussed, both the watchman and early stopping help alleviate this.

\begin{figure}
    \centering
    \includegraphics[scale=0.35]{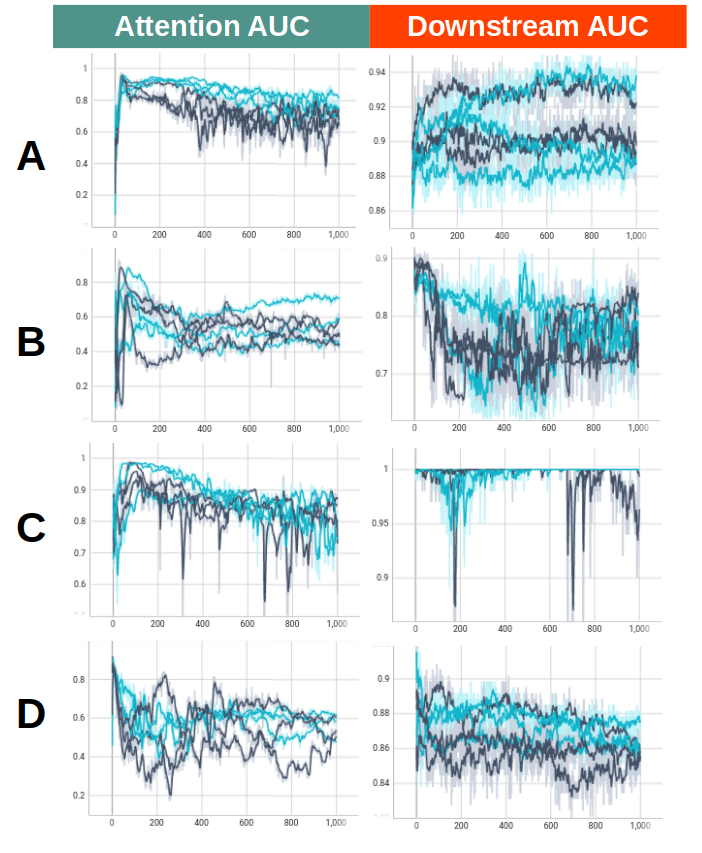}
    \caption{Example of training curves, giving interpretability AUC and downstream accuracy. Light blue is run with Watchman, dark blue without it.
    Those curves are: (A) for GSAT on \textit{Mutag}, (B) for INGENIOUS on \textit{Mutag}, (C) for INGENIOUS on \textit{BA2Motifs}, (D) for the simple double augmentation simclr without regularisation on \textit{Mutag}. The watchman helps prevent overtraining, on both supervised (GSAT) and unsupervised (INGENIOUS). The simple double-aug loss benefits too, showing that any form of regularisation is good. \label{watchman_curves_cherrypicked}}
\end{figure}

Moreover, defining the right number of training epochs is challenging in unsupervised settings. Our training curves (Figure~\ref{watchman_curves_cherrypicked}) show that these unsupervised frameworks are prone to overfitting. This module can attenuate or delay this overfitting.

\newpage
\section{Schematics and Other Figures}

\begin{figure}[h]
    \centering
    \includegraphics[scale=0.31]{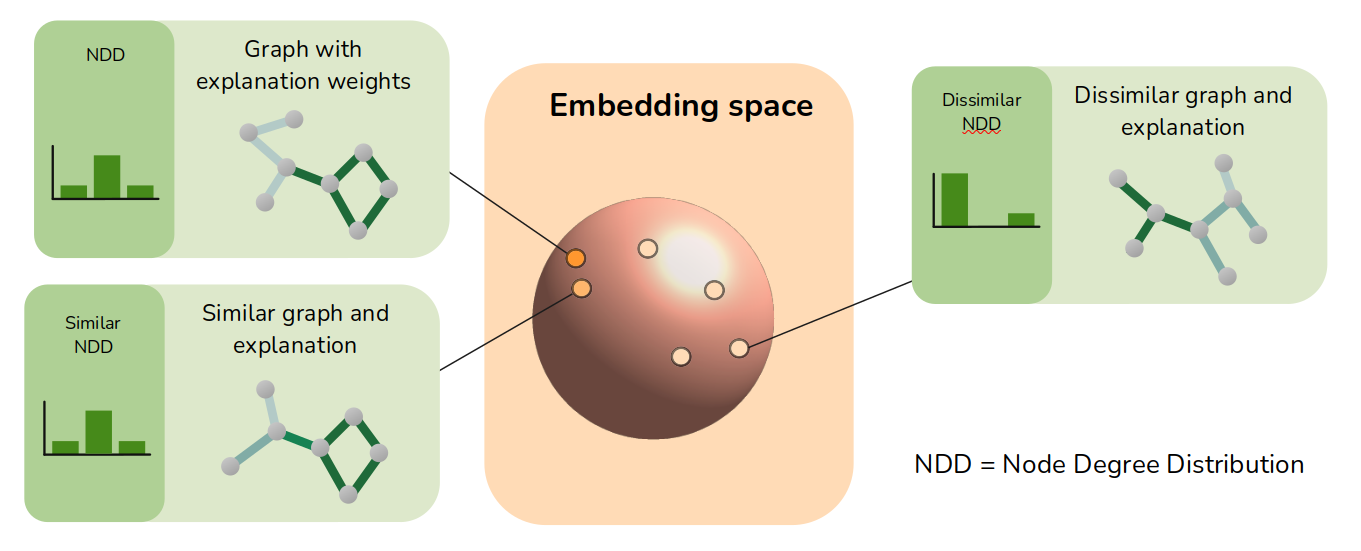}
    \caption{Principle of the Wasserstein distance metric. Our goal is to quantify the continuity of the interpretability functions. Therefore we evaluate whether similar graph embeddings obtain similar interpretations. Graph embedding proximity can be quantified by any vector similarity metric (we used Euclidian distance) and interpretation similarity can be quantified with any graph similarity metric (we use the Wasserstein distance between degree distributions).}
    \label{temperature}
\end{figure}

\begin{figure}{
    \centering
    \includegraphics[scale=0.3]{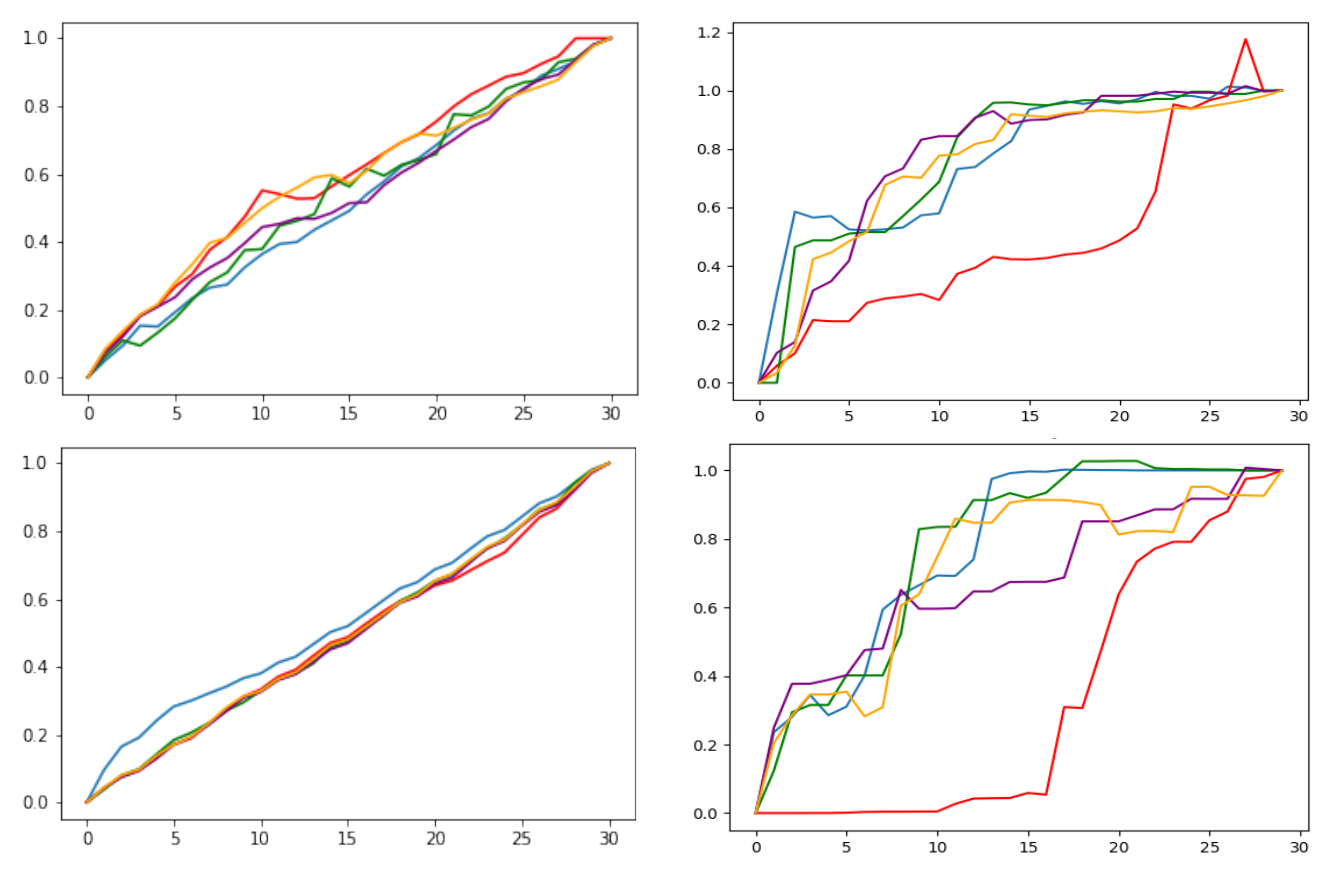}
    \caption{Examples of faithfulness curves on one graph of the \textit{BA2Motifs} datasets. Clockwise starting from the top left: AD-GCL, GSAT, INGENIOUS, MEGA. We show the faithfulness (blue), faithfulness by removing edges according to ground truth (green), shuffled faithfulness (orange), random faithfulness (purple) and opposite faithfulness (red).}
    \label{faithfulness_examples}
    }
\end{figure}
\newpage
\newpage

\section{Full Tables}

\begingroup\centering
\begin{table}[htp]
\resizebox{\linewidth}{!}{%
\begin{tabular}{llllllll}
\toprule
Wm. &                            Methods & \multicolumn{3}{c}{Faithfulness Gap} & \multicolumn{3}{c}{Wasserstein Gap} \\
  &           &         BA2Motifs &            Mutag &      SPMotifs.5 &                  BA2Motifs &            Mutag &      SPMotifs.5 \\
\midrule
        &  GSAT &  $.34 \pm .00$ &  $.21 \pm .08$ &   $.06 \pm .02$ &  $.04 \pm .00$ &  \uline{$.02 \pm .00$} &  $.04 \pm .00$ \\

        &	AD-GCL&	$.01 \pm .13$	&$.03 \pm .09$	&$.08 \pm .06$	& \uline{$.10 \pm .01$}	&$.01 \pm .00$	&$.03 \pm .00$ \\
        &   MEGA&	$.06 \pm .02$	&$.39 \pm .12$	& \uline{$.68 \pm .03$}	&$.04 \pm .01$	&$.01 \pm .00$	&$.01 \pm .00$ \\

        &  INGENIOUS &  $.39 \pm .05$ &  $.24 \pm .01$ &     $.50 \pm 0.00$ &  $.06 \pm .00$ &  $.01 \pm .00$ &    \uline{$.06 \pm 0.00$} \\
         & INGENIOUS - Info & $ .54 \pm .03$ &  \uline{$.58 \pm .04$} &   $.46 \pm .07$ &  $.02 \pm .00$ &  $.01 \pm .00$ &  $.05 \pm .01$ \\
        & INGENIOUS - Negative &  $.13 \pm .02$ &  $.19 \pm .02$ &   $.23 \pm .04$ &  $.02 \pm .00$ &  $.00 \pm .00$ &  $.02 \pm .00$ \\
      
        &  INGENIOUS - Negative - Info&  $.58 \pm .13$ &  $.17 \pm .13$ &   $.27 \pm .08$ &  $.04 \pm .03$ &  $.00 \pm .00$ &  $.02 \pm .00$ \\

        \checkmark& GSAT &  $.25 \pm .03$ &  $.06 \pm .10$ &  $-.03 \pm .01$ &  $.04 \pm .00$ &  \uline{$.02 \pm .00$} &  $.04 \pm .00$ \\
        \checkmark&  INGENIOUS &  $.60 \pm .02$ &  $.41 \pm .18$ &   $.45 \pm .03$ &  $.06 \pm .01$ &  $.01 \pm .00$ &  $.05 \pm .01$ \\
       \checkmark& INGENIOUS - Info &  \uline{$.68\pm .02$} &  $.48 \pm .03$ &   $.48 \pm .06$ &  $.05 \pm .02$ &  $.01 \pm .00$ &  $.05 \pm .01$ \\
       \checkmark & INGENIOUS - Negative &  $.13 \pm .02$ &  $.18 \pm .00$ &   $.22 \pm .06$ &  $.02 \pm .01$ &  $.01 \pm .00$ &  $.02 \pm .01$ \\
        \checkmark& INGENIOUS - Negative - Info &  $.61 \pm .10$ &  $.15 \pm .09$ &   $.32 \pm .00$ &  $.03 \pm .01$ &  $.01 \pm .00$ &  $.03 \pm .00$ \\

\bottomrule
\end{tabular} } 
 \caption{Table of full results. A check in the Wm column means the model uses the watchman loss. A higher faithfulness gap means the interpretations are more faithful, a higher wasserstein gap means the embedding space is more continuous in terms of interpretations.\label{table_res}}
\end{table}
\endgroup

\begingroup\centering
\begin{table}[htp]
\resizebox{\linewidth}{!}{%
\begin{tabular}{llllllll}
\toprule
Wm. &                            Loss & \multicolumn{3}{c}{Downstream ACC} & \multicolumn{3}{c}{Intepretability AUC} \\
     &        &                   Cora &       Tree-cycle &        Tree-grid &             Cora &       Tree-cycle &        Tree-grid \\
\midrule
        &                            GSAT &       $.610 \pm .07$ & $.985 \pm .01$ & $.984 \pm .01$ & $.000 \pm .00$ & $.630 \pm .17$ & $.844 \pm .01$ \\
        & $\mathcal{L}$ &       $.572 \pm .06$ & $.848 \pm .06$ & $.954 \pm .02$ & $.000 \pm .00$ & $.232 \pm .11$ & $.532 \pm .28$ \\
        &      $\mathcal{L}$ - Info &       $.593 \pm .05$ & $.777 \pm .11$ & $.952 \pm .03$ & $.000 \pm .00$ & $.256 \pm .09$ & $.588 \pm .34$ \\
        &          $\mathcal{L}$ - Negative &       $.575 \pm .06$ & $.951 \pm .01$ & $.868 \pm .08$ & $.000 \pm .00$ & $.293 \pm .01$ & $.549 \pm .04$ \\
        &               $\mathcal{L}$ - Negative - Info &       $.606 \pm .04$ & $.856 \pm .02$ & $.887 \pm .07$ & $.000 \pm .00$ & $.319 \pm .08$ & $.702 \pm .07$ \\
\bottomrule

\end{tabular}
}\caption{Downstream ACC and interpretability AUC on node classification. Higher is better\label{donwstream_nodes}} 

\end{table}
\endgroup

\begingroup\centering
\begin{table}
\resizebox{\linewidth}{!}{%
\begin{tabular}{llllllll}
\toprule
Wm. &                            Loss & \multicolumn{3}{c}{Downstream ACC} & \multicolumn{3}{c}{Intepretability AUC} \\
     &        &             BA2Motifs &            Mutag &      SPMotifs.5 &       BA2Motifs &            Mutag &      SPMotifs.5 \\
\midrule
        &                            GSAT &       $1 \pm .00$ & $.921 \pm .02$ & $.393 \pm .02$ & $.998 \pm .00$ & $.843 \pm .15$ & $.895 \pm .01$ \\
        &                           AD-GCL &       $1 \pm .00$ & $.902 \pm .02$ & $.337 \pm .00$ & $.378 \pm .06$ & $.416 \pm .19$ & $.472 \pm .03$ \\
        &                            MEGA &       $1 \pm .00$ & $.886 \pm .01$ & $.335 \pm .00$ & $.459 \pm .28$ & $.566 \pm .46$ & $.506 \pm .03$ \\
        & $\mathcal{L}$ &       $1 \pm .00$ & $.900 \pm .01$ &  $.366 \pm .01$ & $.910 \pm .04$ & $.745 \pm .11$ &  $.467 \pm .01$ \\
        &      $\mathcal{L}$ - Info &       $.990 \pm .00$ & $.889 \pm .03$ & $.361 \pm .02$ & $.860 \pm .07$ & $.606 \pm .23$ & $.491 \pm .10$ \\
        &          $\mathcal{L}$ - Negative &       $1 \pm .00$ & $.887 \pm .01$ & $.340 \pm .02$ & $.745 \pm .24$ & $.546 \pm .04$ & $.494 \pm .02$ \\
        &               $\mathcal{L}$ - Negative - Info &       $1 \pm .00$ & $.895 \pm .01$ & $.337 \pm .01$ & $.462 \pm .38$ & $.478 \pm .23$ & $.494 \pm .04$ \\
        \checkmark &                            GSAT &       $1 \pm .00$ & $.912 \pm .00$ & $.389 \pm .01$ & $.999 \pm .00$ & $.904 \pm .04$ & $.882 \pm .03$ \\
        \checkmark & $\mathcal{L}$ &       $.993 \pm .01$ & $.900 \pm .03$ & $.363 \pm .00$ & $.959 \pm .01$ & $.771 \pm .16$ & $.510 \pm .06$ \\
        \checkmark &      $\mathcal{L}$ - Info &       $1 \pm .00$ & $.863 \pm .04$ & $.360 \pm .02$ & $.900 \pm .02$ & $.563 \pm .20$ & $.499 \pm .07$ \\
        \checkmark &          $\mathcal{L}$ - Negative &       $.995 \pm .01$ & $.895 \pm .01$ & $.338 \pm .00$ & $.148 \pm .00$ & $.620 \pm .03$ & $.581 \pm .03$ \\
        \checkmark &               $\mathcal{L}$ - Negative - Info &       $.997 \pm .01$ & $.903 \pm .01$ & $.337 \pm .00$ & $.583 \pm .30$ & $.650 \pm .18$ & $.498 \pm .04$ \\
\bottomrule
\end{tabular}
}
\caption{Downstream ACC and interpretability AUC on graph classification Higher is better.\label{downstream_graph}}
\end{table}
\endgroup

\begingroup\centering
\begin{table}
\resizebox{\linewidth}{!}{%
\begin{tabular}{llllllll}
\toprule
Wm. &                            Loss & \multicolumn{3}{c}{Random faithfulness} & \multicolumn{3}{c}{Sparsity} \\
   &          &                   BA2Motifs &            Mutag &      SPMotifs.5 &         BA2Motifs &            Mutag &      SPMotifs.5 \\
\midrule
        &                            GSAT &             $.840 \pm .04$ & $.731 \pm .03$ & $.708 \pm .07$ &   $.883 \pm .01$ & $.812 \pm .14$ & $.874 \pm .00$ \\
        &                           AD-GCL &             $.606 \pm .02$ & $.774 \pm .03$ & $.679 \pm .01$ &   $.608 \pm .08$ & $.955 \pm .05$ & $1 \pm .00$ \\
        &                            MEGA &             $.486 \pm .01$ & $.638 \pm .02$ & $.614 \pm .01$ &   $.566 \pm .13$ & $.549 \pm .07$ & $.275 \pm .05$ \\
        & $\mathcal{L}$ &             $.875 \pm .02$ & $.804 \pm .03$ &  $.649 \pm .01$ &   $.523 \pm .03$ & $.897 \pm .11$ &  $.180 \pm .01$ \\
        &      $\mathcal{L}$ - Info &             $.714 \pm .09$ & $.666 \pm .08$ & $.651 \pm .03$ &   $.266 \pm .12$ & $.191 \pm .20$ & $.166 \pm .01$ \\
        &          $\mathcal{L}$ - Negative &             $1.020 \pm .04$ & $.859 \pm .01$ & $.647 \pm .06$ &   $.991 \pm .00$ & $.980 \pm .00$ & $.714 \pm .24$ \\
        &               $\mathcal{L}$ - Negative - Info &             $.897 \pm .13$ & $.879 \pm .02$ & $.641 \pm .06$ &   $.417 \pm .13$ & $.958 \pm .07$ & $.657 \pm .29$ \\
        \checkmark &                            GSAT &             $.817 \pm .02$ & $.822 \pm .01$ & $.538 \pm .05$ &   $.855 \pm .04$ & $.893 \pm .00$ & $.848 \pm .00$ \\
        \checkmark & $\mathcal{L}$ &             $.719 \pm .01$ & $.783 \pm .02$ & $.636 \pm .01$ &   $.402 \pm .04$ & $.591 \pm .31$ & $.205 \pm .06$ \\
        \checkmark &      $\mathcal{L}$ - Info &             $.743 \pm .03$ & $.670 \pm .09$ & $.606 \pm .02$ &   $.288 \pm .04$ & $.269 \pm .23$ & $.168 \pm .01$ \\
        \checkmark &          $\mathcal{L}$ - Negative &             $.877 \pm .03$ & $.808 \pm .02$ & $.663 \pm .07$ &   $.840 \pm .02$ & $.969 \pm .00$ & $.808 \pm .31$ \\
        \checkmark &               $\mathcal{L}$ - Negative - Info &             $.767 \pm .10$ & $.789 \pm .05$ & $.594 \pm .02$ &   $.402 \pm .09$ & $.971 \pm .04$ & $.466 \pm .10$ \\
\bottomrule
\end{tabular}}
\caption{Random faithfulness and sparsity on graph classification. For random faithfulness, higher is better and for sparsity lower is better.\label{downstream_nodes} }
\end{table}
\endgroup

\begingroup\centering
\begin{table}
\resizebox{\linewidth}{!}{%
\begin{tabular}{llllllll}
\toprule
Wm. &                            Loss & \multicolumn{3}{c}{ faithfulness} & \multicolumn{3}{c}{Opposite faithfulness} \\
     &        &               Cora &       Tree-cycle &        Tree-grid &                        Cora &       Tree-cycle &        Tree-grid \\
\midrule
        &                            GSAT &   $.756 \pm .01$ & $.728 \pm .21$ & $.960 \pm .03$ &            $.550 \pm .07$ & $.761 \pm .05$ & $.691 \pm .01$ \\
        & $\mathcal{L}$ &   $.747 \pm .03$ & $.839 \pm .07$ & $.965 \pm .07$ &            $.628 \pm .13$ & $.699 \pm .08$ & $.707 \pm .05$ \\
        &      $\mathcal{L}$ - Info &   $.752 \pm .03$ & $.767 \pm .03$ & $.868 \pm .06$ &            $.611 \pm .10$ & $.680 \pm .17$ & $.584 \pm .11$ \\
        &          $\mathcal{L}$ - Negative &   $.732 \pm .01$ & $.853 \pm .08$ & $.902 \pm .03$ &            $.531 \pm .06$ & $.846 \pm .05$ & $.610 \pm .15$ \\
        &               $\mathcal{L}$ - Negative - Info &   $.783 \pm .04$ & $.780 \pm .04$ & $.847 \pm .01$ &            $.477 \pm .03$ & $.608 \pm .05$ & $.598 \pm .11$ \\
\bottomrule
\end{tabular}}
\caption{Faithfulness and opposite faithfulness on node classification. For faithfulness, higher is better and for opposite faithfulness lower is better.\label{full_faithfulness_nodes}}
\end{table}
\endgroup

\begingroup\centering
\begin{table}
\resizebox{\linewidth}{!}{%
\begin{tabular}{llllllll}
\toprule
Wm. &                            Loss & \multicolumn{3}{c}{Random faithfulness} & \multicolumn{3}{c}{Sparsity} \\
     &        &                         Cora &       Tree-cycle &        Tree-grid &               Cora &       Tree-cycle &        Tree-grid \\
\midrule
        &                            GSAT &             $.693 \pm .06$ & $.814 \pm .12$ & $.911 \pm .02$ &   $.708 \pm .01$ & $.827 \pm .04$ & $.801 \pm .00$ \\
        & $\mathcal{L}$ &             $.727 \pm .02$ & $.823 \pm .13$ & $.877 \pm .04$ &   $.754 \pm .06$ & $.629 \pm .08$ & $.639 \pm .06$ \\
        &      $\mathcal{L}$ - Info &             $.727 \pm .03$ & $.757 \pm .07$ & $.816 \pm .10$ &   $.711 \pm .04$ & $.481 \pm .03$ & $.565 \pm .11$ \\
        &          $\mathcal{L}$ - Negative &             $.718 \pm .03$ & $.818 \pm .08$ & $.849 \pm .01$ &   $.669 \pm .04$ & $.827 \pm .01$ & $.609 \pm .12$ \\
        &               $\mathcal{L}$ - Negative - Info &             $.714 \pm .04$ & $.735 \pm .06$ & $.816 \pm .03$ &   $.740 \pm .01$ & $.564 \pm .10$ & $.592 \pm .05$ \\
\bottomrule
\end{tabular}}
\caption{Random Faithfulness and Sparsity on node classification. For random faithfulness, higher is better and for sparsity lower is better.\label{random_sparsity_nodes}}
\end{table}
\endgroup

\begingroup\centering
\begin{table}[htp]
\resizebox{\linewidth}{!}{%
\begin{tabular}{llllllll}
\toprule
Wm. &                            Method & \multicolumn{3}{c}{$W_1$ global} & \multicolumn{3}{c}{$W_1$ local} \\
    &         &                     BA2Motifs &            Mutag &      SPMotifs.5 &                           BA2Motifs &            Mutag &      SPMotifs.5 \\
\midrule
        &                            GSAT &               $.065 \pm .02$ & $.097 \pm .00$ & $.080 \pm .00$ &                     $.019 \pm .01$ & $.071 \pm .01$ & $.036 \pm .00$ \\
        &                           AD-GCL &               $.240 \pm .04$ & $.099 \pm .00$ & $.117 \pm .03$ &                     $.134 \pm .02$ & $.085 \pm .00$ & $.082 \pm .02$ \\
        &                            MEGA &               $.139 \pm .04$ & $.128 \pm .01$ & $.115 \pm .00$ &                     $.095 \pm .03$ & $.113 \pm .01$ & $.099 \pm .01$ \\
        & INGENIOUS &               $.154 \pm .01$ & $.099 \pm .00$ &  $.113 \pm .01$ &                     $.093 \pm .02$ & $.087 \pm .01$ &  $.044 \pm .01$ \\
        &      INGENIOUS - Info &               $.144 \pm .01$ & $.096 \pm .01$ & $.105 \pm .02$ &                     $.122 \pm .01$ & $.084 \pm .00$ & $.047 \pm .01$ \\
        &          INGENIOUS - Negative &               $.123 \pm .01$ & $.100 \pm .00$ & $.132 \pm .01$ &                     $.101 \pm .01$ & $.095 \pm .00$ & $.103 \pm .01$ \\
        &               INGENIOUS - Negative - Info &               $.144 \pm .02$ & $.100 \pm .00$ & $.117 \pm .01$ &                     $.098 \pm .02$ & $.092 \pm .00$ & $.092 \pm .02$ \\
        \checkmark &                            GSAT &               $.055 \pm .01$ & $.097 \pm .00$ & $.080 \pm .00$ &                     $.012 \pm .00$ & $.077 \pm .00$ & $.035 \pm .00$ \\
        \checkmark & INGENIOUS &               $.143 \pm .01$ & $.099 \pm .00$ & $.095 \pm .01$ &                     $.080 \pm .01$ & $.084 \pm .00$ & $.044 \pm .00$ \\
        \checkmark &      INGENIOUS - Info &               $.151 \pm .01$ & $.100 \pm .00$ & $.094 \pm .02$ &                     $.098 \pm .01$ & $.087 \pm .00$ & $.042 \pm .00$ \\
        \checkmark &          INGENIOUS - Negative &               $.136 \pm .00$ & $.101 \pm .00$ & $.116 \pm .01$ &                     $.107 \pm .02$ & $.089 \pm .00$ & $.088 \pm .02$ \\
        \checkmark &               INGENIOUS - Negative - Info &               $.140 \pm .00$ & $.101 \pm .00$ & $.110 \pm .01$ &                     $.105 \pm .01$ & $.087 \pm .00$ & $.074 \pm .01$ \\
\bottomrule
\end{tabular} }
\caption{Global and local Wasserstein distances on graph embeddings.\label{wasserstein_full_table}}
\end{table}
\endgroup

\begingroup\centering
\begin{table}[htp]
\resizebox{\linewidth}{!}{%
\begin{tabular}{llllllll}
\toprule
Wm. &                            Method & \multicolumn{3}{c}{ Faithfulness} & \multicolumn{3}{c}{Opposite Faithfulness} \\
  &           &         BA2Motifs &            Mutag &      SPMotifs.5 &                  BA2Motifs &            Mutag &      SPMotifs.5 \\
\midrule
        &                            GSAT &   $.843 \pm .02$ & $.778 \pm .05$ & $.702 \pm .06$ &            $.497 \pm .02$ & $.565 \pm .04$ & $.643 \pm .03$ \\
        &                           AD-GCL &   $.622 \pm .07$ & $.779 \pm .00$ & $.707 \pm .02$ &            $.609 \pm .06$ & $.747 \pm .09$ & $.626 \pm .05$ \\
        &                            MEGA &   $.531 \pm .01$ & $.827 \pm .04$ & $.898 \pm .01$ &            $.467 \pm .02$ & $.436 \pm .09$ & $.214 \pm .02$ \\
        & INGENIOUS &   $.831 \pm .03$ & $.830 \pm .03$ &  $.832 \pm 0.01$ &            $.432 \pm .05$ & $.588 \pm .04$ &  $.325 \pm 0.01$ \\
        &      INGENIOUS - Info &   $.823 \pm .01$ & $.864 \pm .03$ & $.817 \pm .06$ &            $.274 \pm .04$ & $.279 \pm .03$ & $.356 \pm .03$ \\
        &          INGENIOUS - Negative &   $.978 \pm .05$ & $.834 \pm .02$ & $.744 \pm .03$ &            $.844 \pm .03$ & $.637 \pm .00$ & $.506 \pm .07$ \\
        &               INGENIOUS - Negative - Info &   $.912 \pm .04$ & $.838 \pm .04$ & $.746 \pm .01$ &            $.323 \pm .12$ & $.664 \pm .10$ & $.474 \pm .10$ \\
        \checkmark &                            GSAT &   $.782 \pm .02$ & $.747 \pm .07$ & $.511 \pm .04$ &            $.523 \pm .02$ & $.684 \pm .04$ & $.548 \pm .05$ \\
        \checkmark & INGENIOUS &   $.856 \pm .01$ & $.862 \pm .07$ & $.797 \pm .02$ &            $.257 \pm .01$ & $.443 \pm .12$ & $.345 \pm .01$ \\
        \checkmark &      INGENIOUS - Info &   $.884 \pm .02$ & $.819 \pm .07$ & $.796 \pm .02$ &            $.200 \pm .01$ & $.336 \pm .03$ & $.316 \pm .04$ \\
        \checkmark &          INGENIOUS - Negative &   $.801 \pm .00$ & $.803 \pm .01$ & $.738 \pm .02$ &            $.663 \pm .02$ & $.617 \pm .02$ & $.518 \pm .08$ \\
        \checkmark &               INGENIOUS - Negative - Info &   $.893 \pm .05$ & $.782 \pm .03$ & $.730 \pm .01$ &            $.279 \pm .08$ & $.629 \pm .08$ & $.409 \pm .02$ \\
\bottomrule
\end{tabular}
} \caption{Faithfulness depending on experiments. For faithfulness, higher is better, for opposite faithfulness, lower is better.\label{faithfulness_table}} 
\end{table}
\endgroup

\end{document}